\documentclass[conference]{IEEEtran}
\usepackage{times}
\usepackage{amsmath}
\usepackage{booktabs}
\usepackage{graphicx}
\usepackage{wrapfig}
\usepackage{comment}
\usepackage{float}
\usepackage{amsthm}
\usepackage{amssymb}
\usepackage{mathtools}
\usepackage{adjustbox}
\usepackage{float}
\usepackage[normalem]{ulem}
\usepackage{wrapfig}
\usepackage{listings}
\usepackage{color}
\usepackage{multicol}
\usepackage{footnote}
\usepackage[dvipsnames]{xcolor}
\usepackage[small]{caption}
\usepackage[table]{xcolor}
\usepackage{booktabs}
\usepackage{makecell}
\usepackage{pifont}
\usepackage{tcolorbox}
\usepackage{fvextra} %
\newcommand{\cmark}{\ding{51}}

\newcommand{\myparagraph}[1]{\noindent\textbf{#1}~}
\usepackage[numbers,sort&compress]{natbib}
\usepackage{multicol}
\usepackage[bookmarks=true]{hyperref}

\definecolor{headergray}{RGB}{245,245,245}

\newcommand{\projectname}{KinDER}
\newcommand{\gardenname}{KinDERGarden}
\newcommand{\gymname}{KinDERGym}
\newcommand{\benchmarkname}{KinDERBench}

\makeatletter
\apptocmd\@maketitle{{\myfigure{}\par}}{}{}
\newcommand{\removelatexerror}{\let\@latex@error\@gobble}
\makeatother

\begin{document}
\newcommand\myfigure{%
\centering
\noindent
\includegraphics[width=\textwidth]{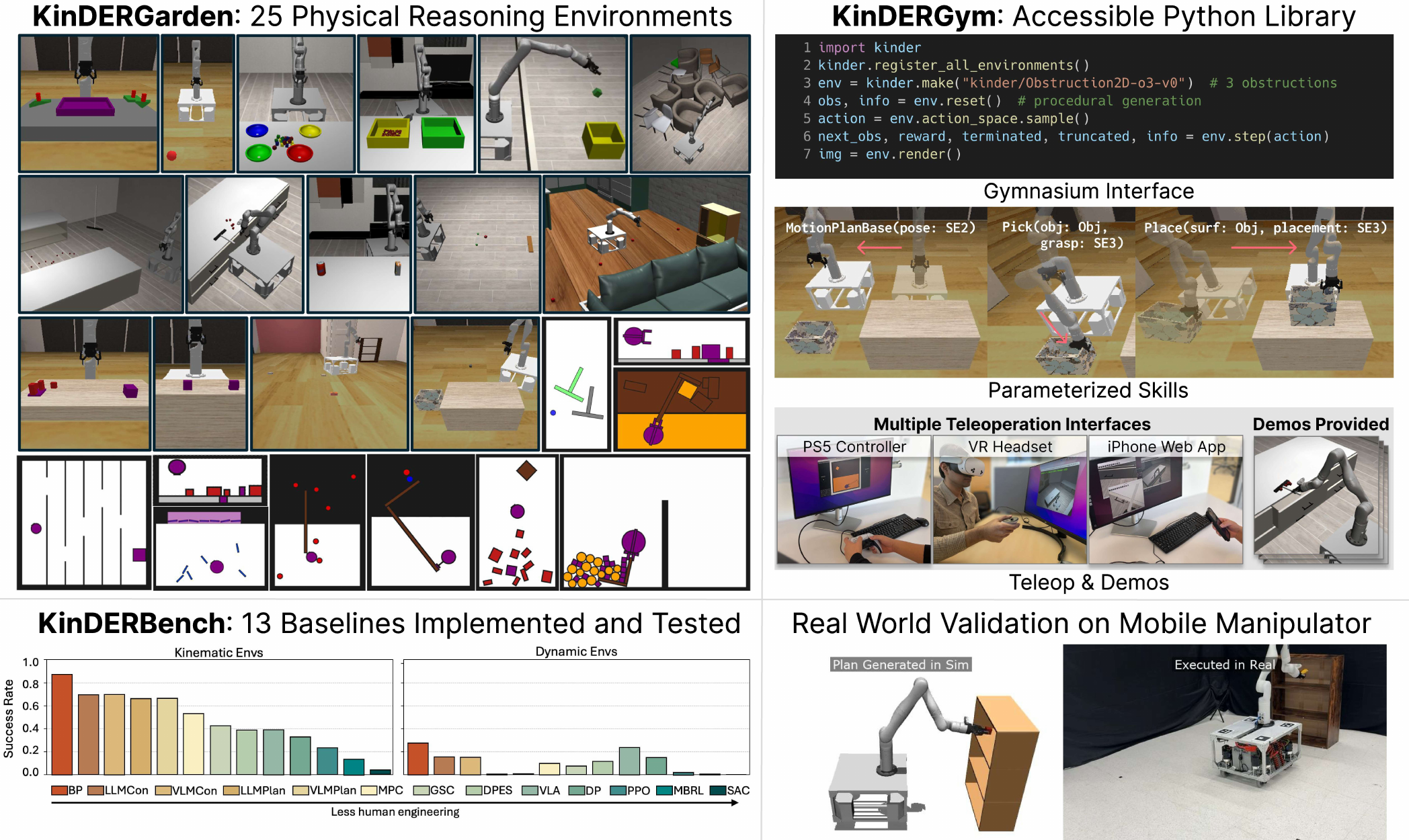}
\captionof{figure}{\textbf{\projectname{} Overview.} We present \textbf{\projectname{}}, a physical reasoning benchmark for robot learning and planning with three main contributions: \textbf{\gardenname{}} (top left), a collection of 25 physical reasoning environments; \textbf{\gymname{}} (top right), a Python library with a Gymnasium interface, parameterized skills, multiple teleoperation interfaces, and demonstrations; and \textbf{\benchmarkname{}} (bottom left), a suite of baselines and evaluations. We also show real-world validity on a mobile manipulator (bottom right).}
\label{fig:teaser}
\vspace{-10pt}
\setcounter{figure}{1}
}

\title{\projectname: A Physical Reasoning Benchmark for Robot Learning and Planning}

\definecolor{flodarkpurple}{rgb}{0.288,0.1196,0.7}
\newcommand{\authorhref}[3][flodarkpurple]{\href{#2}{\textcolor{#1}{#3}}}
\author{%
    \normalfont
    \authorhref{https://yixuanhuang98.github.io/}{Yixuan Huang}\textsuperscript{1*},\,
    \authorhref{https://jaraxxus-me.github.io/}{Bowen Li}\textsuperscript{2*},\,
    \authorhref{https://sites.google.com/view/vaibhavsaxena}{Vaibhav Saxena}\textsuperscript{3*},\,

     \authorhref{https://yichao-liang.github.io/}{Yichao Liang}\textsuperscript{4},\,

     \authorhref{https://umishra.me/}{Utkarsh A. Mishra}\textsuperscript{3},\,

     \authorhref{https://www.researchgate.net/profile/Liang-Ji-13}{Liang Ji}\textsuperscript{1},\, \\ 

     \authorhref{https://lihzha.github.io/}{Lihan Zha}\textsuperscript{1},\,

     \authorhref{https://jimmyyhwu.github.io/}{Jimmy Wu}\textsuperscript{5},\,

     \authorhref{https://nishanthjkumar.com/}{Nishanth Kumar}\textsuperscript{6},\,

     \authorhref{https://www.ri.cmu.edu/ri-faculty/sebastian-scherer/}{Sebastian Scherer}\textsuperscript{2},\,

     \authorhref{https://faculty.cc.gatech.edu/~danfei/}{Danfei Xu}\textsuperscript{3,5},\,
    
    \authorhref{https://tomsilver.github.io/}{Tom Silver}\textsuperscript{1}\\\vspace{-10pt}\\
    \textsuperscript{1}\href{https://www.princeton.edu/}{Princeton University},
    \textsuperscript{2}\href{https://www.cmu.edu/}{Carnegie Mellon University},
    \textsuperscript{3}\href{https://www.gatech.edu/}{Georgia Tech},
    \textsuperscript{4}\href{https://www.cam.ac.uk/}{University of Cambridge},
    \textsuperscript{5}\href{https://www.nvidia.com/en-us/}{NVIDIA},
    \textsuperscript{6}\href{https://www.mit.edu/}{MIT}

}

\maketitle

\def\thefootnote{*}\footnotetext{Equal contribution. Correspondence to \tt yh1542@princeton.edu}\def\thefootnote{\arabic{footnote}}

\begin{abstract}
Robotic systems that interact with the physical world must reason about kinematic and dynamic constraints imposed by their own embodiment, their environment, and the task at hand. We introduce \projectname{}, a benchmark for Kinematic and Dynamic Embodied Reasoning that targets physical reasoning challenges arising in robot learning and planning. \projectname{} comprises 25 procedurally generated environments, a Gymnasium-compatible Python library with parameterized skills and demonstrations, and a standardized evaluation suite with 13 implemented baselines spanning task and motion planning, imitation learning, reinforcement learning, and foundation-model-based approaches. 
The environments are designed to isolate five core physical reasoning challenges: basic spatial relations, nonprehensile multi-object manipulation, tool use, combinatorial geometric constraints, and dynamic constraints, disentangled from perception, language understanding, and application-specific complexity. Empirical evaluation shows that existing methods struggle to solve many of the environments, indicating substantial gaps in current approaches to physical reasoning. We additionally include real-to-sim-to-real experiments on a mobile manipulator to assess the correspondence between simulation and real-world physical interaction. \projectname{} is fully open-sourced and intended to enable systematic comparison across diverse paradigms for advancing physical reasoning in robotics.
Website and code: \url{https://prpl-group.com/kinder-site/}
\end{abstract}

\IEEEpeerreviewmaketitle

\section{Introduction}
\label{sec:intro}

\begin{figure*}[!t]
    \centering
    \includegraphics[width=\textwidth]{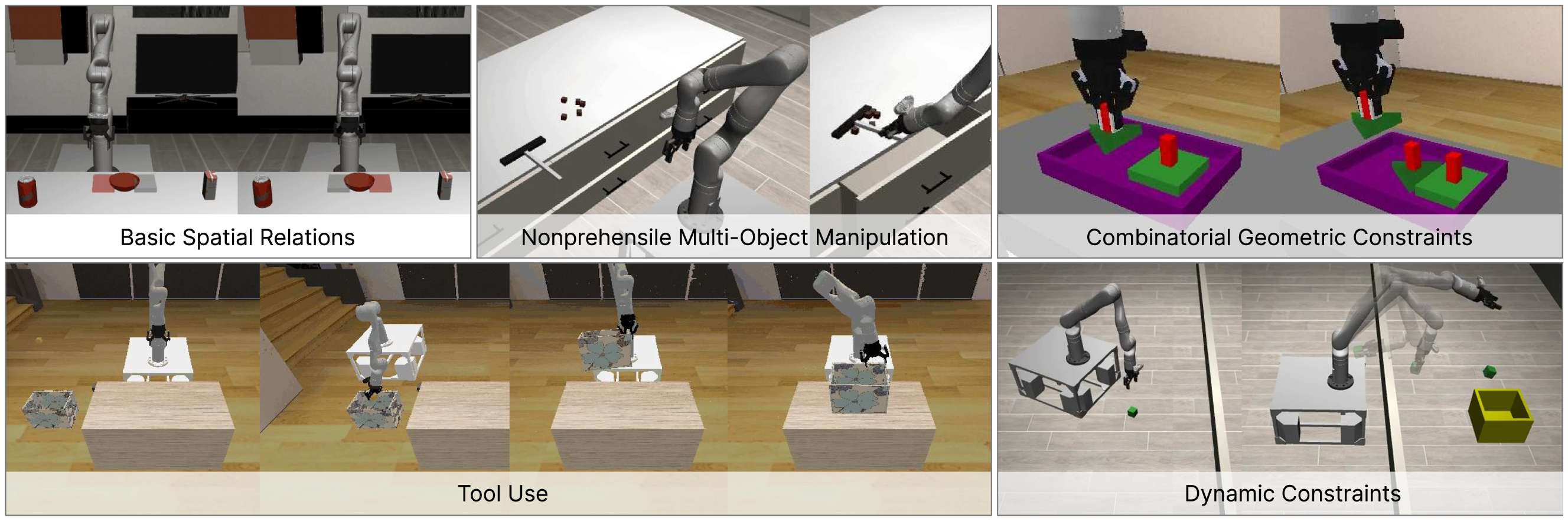}
    \caption{\textbf{Core Challenges for Physical Reasoning in \projectname{}.} From top left: arranging objects to be in goal-specified locations relative to a bowl requires understanding \emph{basic spatial relations}. Sweeping many small objects into a drawer benefits from \emph{nonprehensile multi-object manipulation}. Packing varying numbers of objects into a confined region requires satisfying \emph{combinatorial geometric constraints}. Transporting objects with a box benefits from \emph{tool use}.  Tossing objects over a barrier requires satisfying \emph{dynamic constraints}.}
    \label{fig:core-challenges}
    \vspace{-15pt}
\end{figure*}

A central challenge in robotics that arises from embodied interaction with the real world is the need for \emph{physical reasoning}~\cite{davis2008physical,toussaint2020describing,garrett2021integrated, wang2023newton,cherian2024llmphy,gao2024physically,zhao2025cot,ye2025vla}.
Broadly competent robots must be able to reason about the kinematic and dynamic limits dictated by their own morphology, the laws of physics imposed by their environment, and the task requirements specified by their users.
These often-entangled constraints can turn semantically simple tasks into challenging puzzles~\cite{allen2020rapid, garrett2021integrated,zhao2024survey}.
To store a book in a shelf, is it enough to move to the book, grasp it, move to the shelf, and place?
That depends: is there a clear path to the book, or do obstacles need to be moved?
Does the book need to be set down and re-grasped before a placement is feasible? 
Is there space in the shelf, or should the robot use its arm to gently push other books aside?
The robot must not only answer these questions, but pose them in the first place---\emph{reasoning about what to reason about}~\cite{griffiths2019doing,sung2021learning,vats2022efficient} given only the available sensory observations.

Despite the importance of physical reasoning to robotics, there is little consensus on the state of the art.
Measuring physical reasoning is hard: no single task is sufficient (why not just memorize the solution?) and even procedurally-generated variations~\cite{deitke2022,wang2023gensim,eppner2025scene_synthesizer} of a task cannot capture the challenge of physical reasoning in its full generality.
Existing benchmarks (Section~\ref{sec:related-work}) cover more general challenges for robot learning and planning---broad task diversity, long-horizon decision making, language grounding---or focus on full-fledged application-focused domains such as home assistance.
As a result, it remains difficult to perform targeted evaluation of physical reasoning itself, disentangled from perception, language understanding, or domain-specific considerations.

\begin{table*}[!t]
\centering
\rowcolors{2}{gray!8}{white}
\setlength{\tabcolsep}{10pt}
\renewcommand{\arraystretch}{1.25}

\begin{tabular}{lccccccc}
\toprule
\rowcolor{blue!10}
\textbf{Benchmark} & \textbf{\makecell{Focus On\\Physical\\Reasoning}} & \textbf{2D/3D} & \textbf{\makecell{Basic\\Spatial\\Relations}} & \textbf{\makecell{Nonprehensile\\Multi-Object\\Manipulation}} & \textbf{Tool Use} & \textbf{\makecell{Combinatorial\\Geometric\\Constraints}}  & \textbf{\makecell{Dynamic\\Constraints}} \\
\midrule
BEHAVIOR-1k~\cite{li2024behavior1k} & & 3D & \cmark &  &  \cmark & \cmark  & \cmark  \\ 
CALVIN~\cite{mees2021calvin} & & 3D & \cmark &  &   &  &  \\ 
DittoGym~\cite{li2024dittogym} & \cmark & 2D &  & \cmark  &   &  &  \\ 
DM Control~\cite{tassa2018deepmind} & \cmark & 3D & &  &  &  & \cmark  \\ 
Embodied Interface~\cite{li2024embodied} & & 3D & \cmark &  & \cmark  &  &   \\ 
EmbodiedBench~\cite{yang2025embodiedbench} & & 3D & \cmark &  & \cmark  &  &   \\ 
FurnitureBench~\cite{heo2023furniturebench} & & 3D &  &  & \cmark  & \cmark  &   \\ 
I-PHYRE~\cite{li2024iphyre} & \cmark &   2D & &  & \cmark  &   &  \\
Kinetix~\cite{matthews2024kinetix_openreview} & & 2D &  &  & \cmark  &   & \cmark   \\
\citet{lagriffoul2018platform} & \cmark  & Both &  &  & \cmark  &  & \\ 
LIBERO~\cite{li2023libero} & \cmark  & 3D & \cmark &  &  & \cmark  &  \\
ManiSkill-HAB~\cite{shukla2025maniskillhab} & & 3D & \cmark &  &  & \cmark  &  \\
OGBench~\cite{park2025ogbench} &  & Both &  &  &  &  &  \cmark \\
RoboCasa~\cite{nasiriany2024robocasa}  &   & 3D & &  &  & \cmark  &  \cmark \\
Virtual Tools~\cite{allen2020rapid} & \cmark & 2D &   &  & \cmark  &  & \cmark  \\
VLABench~\cite{zhang2025vlabench} & & 3D & \cmark   &  & \cmark  & \cmark  &  \\
VLMgineer~\cite{gao2025vlmgineer} & \cmark  & 3D &    &  \cmark  & \cmark  &   & \cmark   \\
\textbf{\projectname{} (Ours)} & \cmark  & Both & \cmark & \cmark  & \cmark  & \cmark & \cmark  \\
\bottomrule
\end{tabular}

\caption{\textbf{Related Benchmarks.} \projectname{} is a benchmark for robot physical reasoning with both 2D and 3D environments and with an emphasis on five core challenges (Section~\ref{sec:core-challenges}).
\projectname{} fills a gap in the literature; see Section~\ref{sec:related-work} for discussion.}
\label{tab:related_work}
\vspace{-20pt}
\end{table*}

Another reason for the lack of consensus is that physical reasoning has been studied from very different perspectives in separate subfields of robotics.
Classical approaches such as task and motion planning (TAMP)~\cite{kaelbling2013integrated,srivastava2014combined,garrett2020pddlstream,garrett2021integrated,zhao2024survey} use explicit models and optimization techniques to formulate and solve generalized constraint satisfaction problems.
Model-free approaches such as reinforcement learning (RL)~\cite{schulman2017proximal,haarnoja2018soft,mirza2020physically} and imitation learning (IL)~\cite{osa2018algorithmic,jang2022bc,chi2025diffusion} use data to compile away the need for explicit reasoning.
Foundation model (FM) based approaches such as LLM~\cite{cherian2024llmphy,wang2024llm}, VLM~\cite{gao2024physically,hu2023look}, or VLA~\cite{driess2023palm,zhao2025cot,team2025gemini} planning combine explicit reasoning in natural language with implicit understanding from pretraining.
There is also broad interest in combining the complementary strengths of these approaches, but without clarity on the state of the art, it is difficult to make progress.

To address these challenges, we propose (\textbf{\projectname{}}): a benchmark for \textbf{Kin}ematic and \textbf{D}ynamic \textbf{E}mbodied \textbf{R}easoning.
\projectname{} has three main contributions (Figure~\ref{fig:teaser}):
\begin{enumerate}
    \item \textbf{\gardenname{}}: A collection of 25 simulated environments, each with infinite procedurally-generated variations, to capture different facets of physical reasoning.
    \item \textbf{\gymname{}}: A Python package that includes a Gymnasium-compatible environment API, a collection of parameterized skills and concepts, multiple teleoperation interfaces, and demonstration datasets.
    \item \textbf{\benchmarkname{}}: A standardized benchmark for physical reasoning approaches, with 13 pre-implemented baselines from the literature on TAMP, RL, IL, and FM.
\end{enumerate}
All contributions are open-source and tested on multiple standard operating systems.
To show that the simulated environments map onto real physical reasoning challenges, we additionally report real-to-sim-to-real results.
Taken together, \projectname{} represents a significant step toward clarifying and advancing the state-of-the-art in robot physical reasoning.

\section{\projectname{} Core Challenges}
\label{sec:core-challenges}

We begin by presenting the core physical reasoning challenges that are prioritized in \projectname{}.
To select these challenges, we started by reviewing (1) existing work in robot planning and learning where individual physical reasoning problems are considered with one-off environments; and (2) existing benchmarks in related areas (Section~\ref{sec:related-work}).
We then identified themes in (1) that are not well-represented in (2).
In other words, we chose challenges at the frontier of active research, but where the current state-of-the-art remains unclear.

The five \projectname{} core challenges are illustrated by example in Figure~\ref{fig:core-challenges}.
In Section~\ref{sec:related-work}, we discuss coverage of these challenges in existing benchmarks.
In Section~\ref{sec:kindergarden}, we detail how environments in \gardenname{} capture the challenges.
The challenges are as follows:
\begin{enumerate}
    \item \textbf{Basic Spatial Relations:} To set a dinner table~\cite{gkanatsios2023ebm}, load a dishwasher~\cite{jain2023transformers}, or follow instructions with locative prepositions~\cite{cooper1968semantic}, robots must understand spatial relations between objects. They must have both a passive understanding (is the fork on the left of the plate?) and an active understanding (how can I put it there?).
    \item \textbf{Nonprehensile Multi-Object Manipulation:} Generalized manipulation requires more than pick and place---robots should be able to push~\cite{bicchi1993force}, pull, sweep~\cite{xu2025robopanoptes}, scoop, stir, and slap~\cite{liu2025slap} multiple objects at the same time. They should leverage, rather than strictly avoid, whole-arm~\cite{madan2025prioritouch} and whole-body~\cite{bruedigam2024jacta} contact.
    \item \textbf{Tool Use:} Robots should use objects to manipulate other objects---hammers~\cite{samtani2023learning}, wrenches~\cite{holladay2019force}, hooks~\cite{toussaint2018differentiable}, sticks~\cite{silver23a}, trays~\cite{curtis2022discovering}, bins~\cite{wang2019stable}, and step-stools~\cite{kumar2024practice}. They should understand not only common tool affordances, but also abstract mechanisms so that they can improvise~\cite{nair2019tool}, e.g., use a rock to pitch a tent~\cite{allen2020rapid}.
    \item \textbf{Combinatorial Geometric Constraints:} When object-object and robot-object collisions need to be avoided, e.g., while packing~\cite{nickel2025multi}, retrieving~\cite{nam2021fast}, or navigating around~\cite{stilman2005navigation} objects in tight and cluttered spaces, robots must understand and work within implicit geometric constraints. These constraints are combinatorial~\cite{garrett2021integrated}: when the number of objects grows, the number of constraints (e.g., pairwise collisions) grows polynomially.
    \item \textbf{Dynamic Constraints:} To carry a full cup of coffee~\cite{ichnowski2022gomp}, balance a delicate tray, scoop and pour without spilling~\cite{wang2024grounding}, dribble and toss a basketball~\cite{liu2018learning}, or juggle~\cite{rizzi1992progress}, robots must use control to stabilize dynamical systems. They should understand and obey dynamic constraints, e.g., safety limits on velocity magnitudes or requirements implicit in a task (don't spill).
\end{enumerate}

This list is by no means an exhaustive account of all the challenges associated with robot physical reasoning.
Nonetheless, progress on these challenges would represent a significant step forward for the field.
It is also worth noting that more general decision-making challenges are pervasive in \projectname{}---long task horizons, sparse feedback (goal-based rewards), broad task distributions, and time pressure during planning and execution.
We omit these from the core list above to keep the focus on physical reasoning.

\section{Related Work}
\label{sec:related-work}

We next discuss existing benchmarks and their relationship to \projectname{}.
The main novelty of \projectname{} is its coverage of the core physical reasoning challenges introduced in Section~\ref{sec:core-challenges}.
These challenges are the \emph{focus} in \projectname{}, as opposed to application-driven benchmarks (e.g., home assistance) where physical reasoning is entangled with other factors.
\projectname{} also includes both 2D and 3D environments that permit study of physical reasoning at multiple levels of abstraction.
See Table~\ref{tab:related_work} for a summary of related work.

\paragraph{Benchmarks for Robot Learning and Planning}
There has been a significant amount of work on benchmarking robot learning methods~\cite{li2023libero,mees2021calvin,tassa2018deepmind,li2024embodied,yang2025embodiedbench,heo2023furniturebench,shukla2025maniskillhab,park2025ogbench,nasiriany2024robocasa,zhang2025vlabench,li2024behavior1k,gao2025vlmgineer}.
Some benchmarks are geared toward imitation learning~\cite{mimiclabs} \emph{or} reinforcement learning~\cite{park2025ogbench,tassa2018deepmind,matthews2024kinetix_openreview} \emph{or} foundation-model-based methods~\cite{li2024embodied,yang2025embodiedbench,zhang2025vlabench}; others are explicitly designed to compare different families of techniques.
Table-top manipulation is a common setting~\cite{mees2021calvin,li2023libero,heo2023furniturebench}, but mobile~\cite{shukla2025maniskillhab,nasiriany2024robocasa,li2024behavior1k} and bimanual~\cite{chernyadev2024bigym} manipulation are also considered.
The central technical challenges in these benchmarks include long time horizons, sparse rewards, natural language grounding, and broad task diversity (especially in terms of scene and object variation).
For \projectname{}, we especially take inspiration from LIBERO~\cite{li2023libero} and MimicLabs~\cite{mimiclabs}.

There is far less work on benchmarks for classical robot planning (e.g., task and motion planning).
There are also separate benchmarks for motion planning~\cite{moll2015benchmarking,heiden2021bench,chamzas2021motionbenchmaker} and task planning~\cite{long20033rd, vallati20152014,taitler20242023}.
In particular, the International Planning Competition~\cite{long20033rd, vallati20152014,taitler20242023} has been a longstanding catalyst for task planning research.
To the best of our knowledge, the only benchmark for combined TAMP is the one proposed by~\citet{lagriffoul2018platform}, which is not actively used.
\projectname{} facilitates direct comparisons between robot planning and robot learning methods, and their combinations: \gymname{} provides parameterized skills and concepts, and
\benchmarkname{} reports results for both planning and learning methods.

\paragraph{Application-Driven Benchmarks}
\projectname{} isolates fundamental challenges of physical reasoning so that researchers can get a clear signal as they work on these challenges.
In this sense, \projectname{} is complementary to benchmarks that are explicitly driven by applications.
Home assistance applications are especially well-covered by benchmarks such as ALFRED~\cite{shridhar2020alfred}, AI2-THOR~\cite{kolve2017ai2thor} and ManipulaTHOR~\cite{li2022manipulathor}, BEHAVIOR-1k~\cite{li2024behavior1k}, Habitat~\cite{szot2021habitat}, ManiSkill-HAB~\cite{shukla2025maniskillhab}, and RoboCasa~\cite{nasiriany2024robocasa}.
Other notable and recent application-focused benchmarks include FurnitureBench~\cite{heo2023furniturebench} for furniture assembly, CleanUpBench~\cite{li2025cleanupbench} for sweeping and grasping, and CookBench~\cite{cai2025cookbench} for cooking.
The need for physical reasoning naturally arises in these benchmarks, among many other challenges for robot perception, planning, and learning.
\projectname{} is designed to evaluate and advance robot physical reasoning specifically.

\paragraph{Physical Reasoning Benchmarks}
\projectname{} takes inspiration from benchmarks outside of robotics that focus on physical reasoning such as the Virtual Tools Game~\cite{allen2020rapid} and PHYRE~\cite{bakhtin2019phyre}.
See~\citet{melnik2023benchmarks_tmlr} for a survey.
In contrast to many of these works, our intention is to advance robotics, rather than to better understand human physical reasoning.
Nonetheless, drawing connections between \projectname{} and human-like physical reasoning approaches represents an opportunity for future work~\cite{lake2017building}.
From the perspective of this literature, important aspects of \projectname{} include: continuous spaces, multi-step (long-horizon) decision-making, procedural generation, and kinematic and dynamic constraints.

\begin{figure*}[t]
    \centering
    \includegraphics[width=\textwidth]{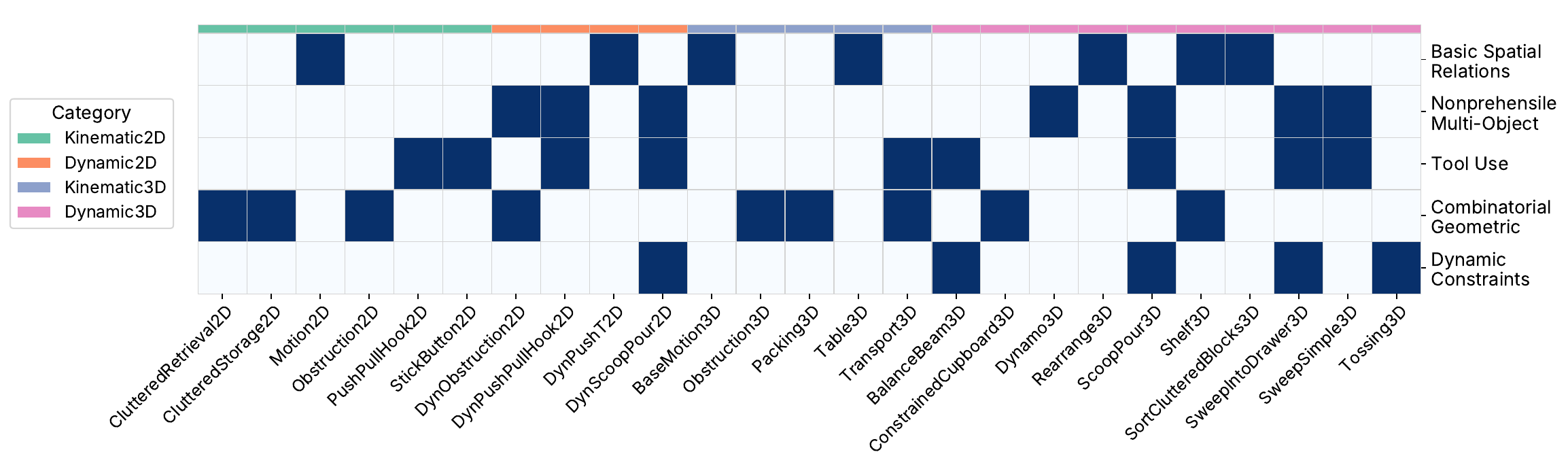}
    \vspace{-20pt}
    \caption{\textbf{\gardenname{} Core Challenges.} Environments in \gardenname{} cover the five core challenges for physical reasoning.}
    \label{fig:challenge-coverage}
    \vspace{-15pt}
\end{figure*}

\begin{figure}[t]
    \centering
    \includegraphics[width=\linewidth]{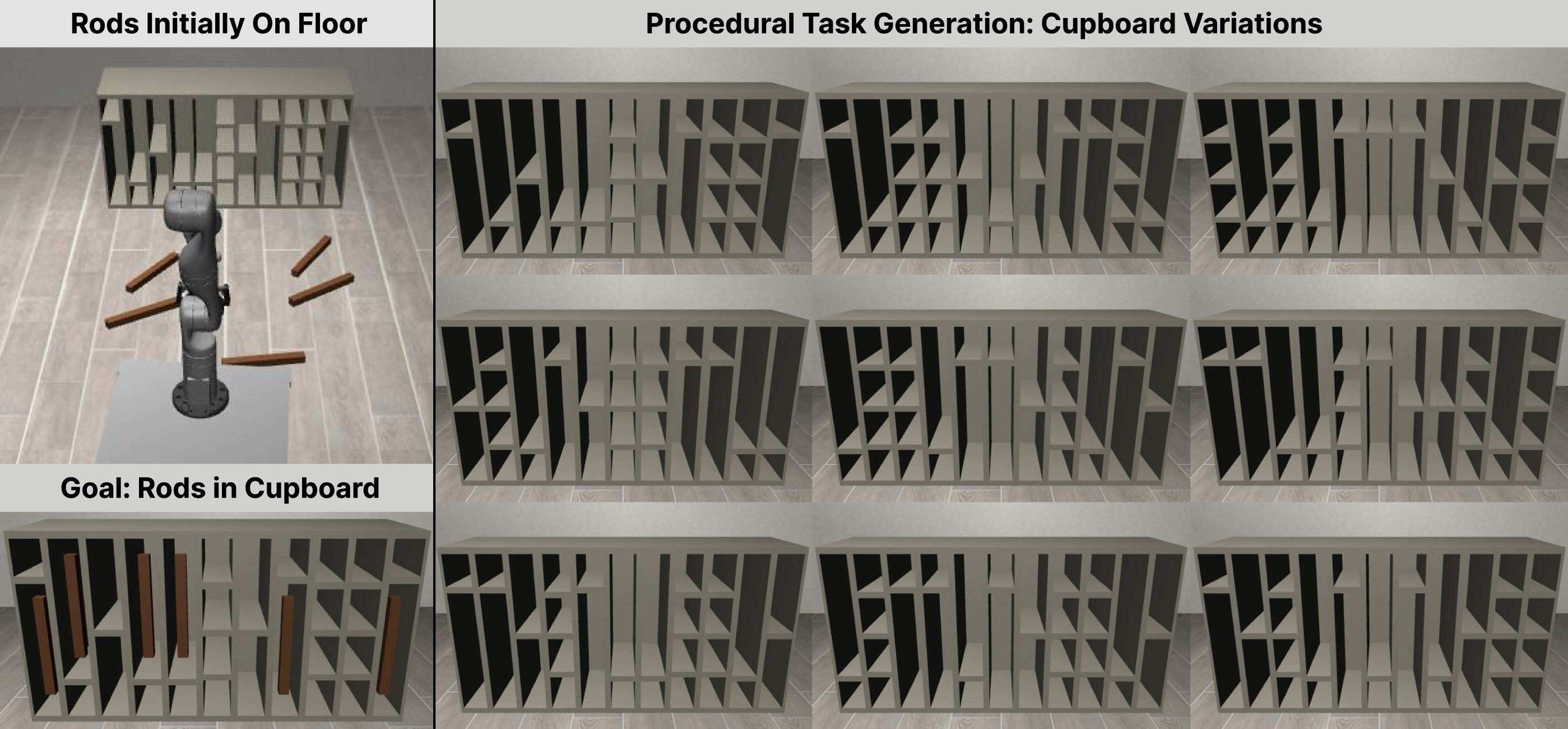}
    \caption{\textbf{Procedural Task Generation Example.} Shelves are randomly arranged within the cupboard in $\mathrm{ConstrainedCupboard3D}$, forcing the robot to reason about the feasibility of rod placements.}
    \label{fig:procedural-generation}
    \vspace{-15pt}
\end{figure}

\section{\gardenname: Environments}
\label{sec:kindergarden}

Our first contribution is \gardenname{}, a collection of 25 environments for robot physical reasoning, grouped into four categories: Kinematic2D, Dynamic2D, Kinematic3D, and Dynamic3D.
We first discuss what is common among all environments and then describe each category.
See Appendix~\ref{app:environments} for details and Figure~\ref{fig:challenge-coverage} for \projectname{} core challenge coverage.

\paragraph{General Environment Structure}

\gardenname{} environments inherit from the general Gymnasium~\cite{towers2024gymnasium} API, which includes an observation space, action space, initial state distribution $\mathrm{reset()}$, and a $\mathrm{step()}$ function that takes an action as input and produces a next observation, reward, and termination indicator.
Rewards are sparse: $-1$ is given at every step until successful termination, which occurs only when a goal is achieved.
All environments have an infinite task distribution that is implemented with procedural generation inside the $\mathrm{reset()}$ function; see Figure~\ref{fig:procedural-generation} for an example.

The main design decision that distinguishes \projectname{} from the general Gymnasium API is that all environments use \emph{object-centric states}.
An object-centric state is a mapping from object names (e.g., $\mathrm{robot}$, $\mathrm{hook}$) to real-valued feature vectors.
The dimensionality of each vector is determined by object \emph{type}.
For example, a $\mathrm{robot}$ with type $\mathrm{MobileManipulator}$ has features for the robot's base position and velocity in $\mathrm{SE(2)}$, arm configuration and velocity in $\mathbb{R}^7$, and gripper joint value in $[0, 1]$.
A $\mathrm{hook}$ with type $\mathrm{Movable}$ has features for pose and velocity in $\mathrm{SE(3)}$ and bounding box dimensions in $\mathbb{R}^3$, among others.
Another $\mathrm{Movable}$ object (e.g., a $\mathrm{plate}$) would have the same feature space.
This design makes it easy to vary the number of objects, which can be useful for evaluating generalization and test-time scaling (Section~\ref{sec:kinderbench}).

Baselines in \projectname{} can use object-centric states directly, but to facilitate experiments with standard learning-based approaches, we provide two other options.
The first option is to use RGB image observations.
The second is to commit to a \emph{variant} of a \gardenname{} environment where the objects are constant.
For example, in $\mathrm{Shelf3D}$, the number of books can vary in general, but in the $\mathrm{b5}$ variant, there are always 5 books.
For constant-object variants, \gardenname{} flattens the object-centric state into a fixed-dimensionality vector.
These environments are then compatible with standard reinforcement learning and imitation learning approaches.

\paragraph{Kinematic2D Environments}

The Kinematic2D category includes six environments that are especially useful for studying tool use and combinatorial geometric constraints at a high level of abstraction.
This category is \emph{kinematic} in the sense that environment transitions are entirely determined by object poses and robot configurations (velocities and accelerations are not modeled); and \emph{2D} in that it is implemented with 2D shapes.
All environments have a robot with a circular base that moves in $\mathrm{SE(2)}$, an extendable 1D arm, and a rectangular vacuum on its end effector that can be activated or deactivated.
When the vacuum is activated, all objects in its immediate vicinity become rigidly attached to the robot.
Actions are constrained to make small changes to the robot's configuration.
When an action is received, a tentative next state is computed.
If that next state includes any collisions, the state is reverted.
These environments are implemented in pure Python; no physics backend is used.

\paragraph{Dynamic2D Environments}

The Dynamic2D category includes four environments that are especially useful for studying nonprehensile multi-object manipulation and tool use at a high level of abstraction.
Unlike Kinematic2D, velocities and accelerations are modeled in this category.
We use the Pymunk physics backend for dynamics~\cite{pymunk}.
Similar to Kinematic2D, these environments feature a robot with a circular base and an extendable 1D arm.
For the benefit of studying contact-rich dynamics, we use a two-fingered gripper on the end effector.

\begin{figure}[t]
    \centering
    \includegraphics[width=\linewidth]{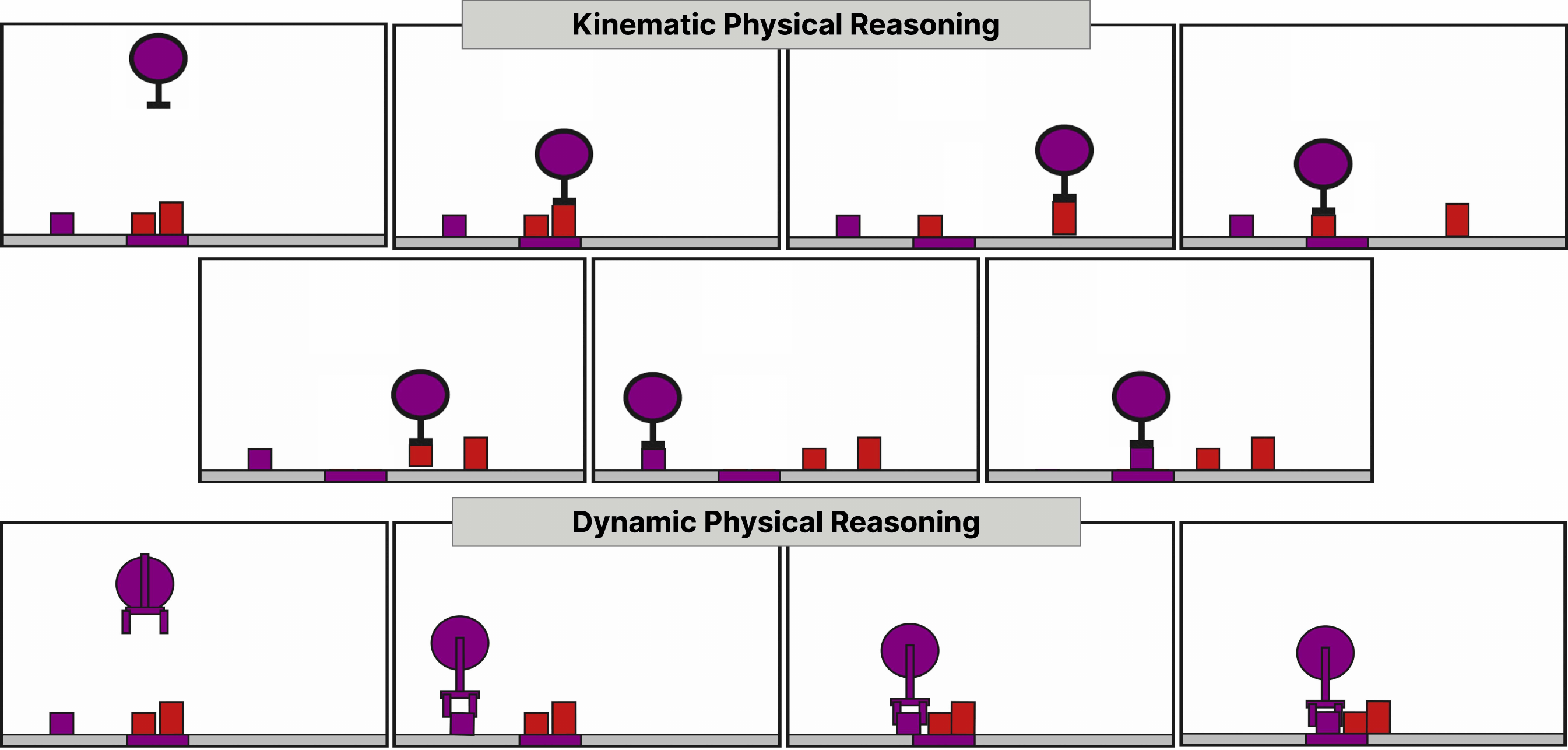}
    \caption{\textbf{2D Kinematic and Dynamic Physical Reasoning Examples.} In $\mathrm{Obstruction2D}$, the robot must pick and place obstacles to make space on a target region. In $\mathrm{DynObstruction2D}$, the robot can push the obstacles out of the way while grasping the target object.}
    \label{fig:kin-vs-dyn-reasoning}
    \vspace{-20pt}
\end{figure}

Kinematic2D and Dynamic2D environments require qualitatively different forms of physical reasoning (Figure~\ref{fig:kin-vs-dyn-reasoning}).
For example, consider the contrast between $\mathrm{Obstruction2D}$ (kinematic) and $\mathrm{DynObstruction2D}$ (dynamic).
In both environments, the goal is to move a target object onto a target region that may be initially obstructed by one or more obstacles.
In the kinematic version, the robot has no choice but to pick and place the obstacles before picking and placing the target object.
However, in the dynamic version, shortcuts~\cite{liu2025slap} are possible: if space constraints allow, the robot may be able to push the obstacles out of the way while holding the target.

\paragraph{Kinematic3D Environments}

The Kinematic3D category includes five environments that are especially useful for studying spatial relations and combinatorial geometric constraints.
These environments are kinematic in the same sense as Kinematic2D (no velocities or accelerations).
We use object modeling, forward kinematics, and collision-checking methods from PyBullet~\cite{coumans2015bullet} to implement transitions in this environment.
For consistency, all environments feature a TidyBot++~\cite{wu2024tidybot++} mobile base with a 7DOF Kinova Gen3 arm~\cite{kinova} and a Robotiq 2F-85~\cite{robotiq} gripper.
When the gripper is closed, objects between the fingers become rigidly attached to the robot until the gripper is opened.
As with Kinematic2D, actions are constrained to make small changes to the robot's configuration; states are reverted when collisions are detected.

\paragraph{Dynamic3D Environments}

The Dynamic3D category includes 10 environments that collectively cover all five core physical reasoning challenges.
Velocities and accelerations are modeled; we use the MuJoCo physics backend for dynamics~\cite{todorov2012mujoco}.
For consistency, we use the same TidyBot++ mobile manipulator as in Kinematic3D.
Unlike Kinematic3D, grasping is dynamic---objects are never rigidly attached to the robot.
Inspired by other MuJoCo-based benchmarks such as LIBERO~\cite{li2023libero}, we use environment configuration files so that all Dynamic3D environments share the same Python code and differ only in their configurations.
We also take inspiration from the BDDL specification language introduced in BEHAVIOR~\cite{li2024behavior1k} in our implementation of procedural task generation, and leverage object and scene assets from RoboCasa~\cite{nasiriany2024robocasa} and MimicLabs~\cite{mimiclabs} respectively.

\section{\gymname: Accessible Software}
\label{sec:kindergym}

Our second main contribution is \gymname{}, a pip-installable Python package that includes not only an interface to the environments in \gardenname{}, but also (1) parameterized skills and concepts; (2) teleoperation interfaces; and (3) precollected demonstrations.
To facilitate ease of use, we developed \gymname{} following strict software engineering standards including continuous integration, linting, type checking, autoformatting, and nearly 400 unit tests.
We have tested Python versions 3.10, 3.11, 3.12, Ubuntu 20.04, 22.04, and 24.04, and macOS 12-15, and Windows 10.

\paragraph{Parameterized Skills and Concepts}

\gymname{} provides utilities for defining parameterized skills and concepts that can be used for hierarchical planning and learning.
Skills are implemented as options~\cite{sutton1999between} with associated PDDL operators~\cite{McDermott1998PDDL} and samplers~\cite{garrett2020pddlstream}.
The options have both object parameters (the same as the PDDL operator) and additional parameters of any type (proposed by the sampler).
For example, a $\mathrm{Pick(object, \theta)}$ skill can be used to pick different objects with different relative grasps $\theta \in \mathrm{SE(3)}$.
For generality, we allow option policies to maintain internal state.
A common pattern is to generate and follow a motion plan.

Concepts are implemented as relational predicates with classifiers that ground in object-centric states~\cite{silver2023predicateinvent,li2025IVNTR}.
For example, $\mathrm{On(object, surface)}$ is a predicate with a classifier that evaluates to True in states where the $\mathrm{object}$ is above and in contact with the $\mathrm{surface}$.
These predicates are used in the preconditions and effects of the skill operators.
Together with the object-centric states, concepts can also be understood as defining a two-level scene graph~\cite{agia2022taskography}.

In our experiments (Section~\ref{sec:kinderbench}), we use \gymname{} skills and concepts for the bilevel planning, LLM planning, and VLM planning baselines.
However, the nature of physical reasoning is such that hierarchical task decompositions are not always readily apparent or easy to engineer.
Designing or learning such skills remains an important direction for future work on physical reasoning that \projectname{} can support.

\paragraph{Teleoperation Interfaces and Demonstrations}

\gymname{} includes multiple teleoperation interfaces that can be used to collect human demonstrations.
Kinematic2D and Dynamic2D environments can be controlled through a mouse-and-keyboard interface, or through a PS5 video game controller.
The mouse-and-keyboard interface includes joystick-like buttons that can be clicked and dragged to move the robot in $\mathrm{SE(2)}$.
Keyboard commands extend and retract the arm, activate and deactivate the vacuum (for Kinematic2D), and open and close the gripper (for Dynamic2D).
The PS5 controller similarly uses the joysticks to move the robot and buttons for the arm, vacuum, and gripper.

Kinematic3D and Dynamic3D environments can be controlled through an iPhone web app, or through a Meta Quest 3S virtual reality headset and controller.
The iPhone web app is based on the TidyBot++ interface~\cite{wu2024tidybot++}, which uses the iPhone's gyroscope and accelerometer to capture spatial inputs.
The teleoperator can toggle between base and arm control.
For arm control, inputs are mapped to task (end effector) space and inverse kinematics is used to derive environment actions.
The Meta Quest 3S interface uses the right-hand controller for task-space inputs and the left-hand controller for base movements.
We additionally allow the teleoperator to select one or more camera angles, which can also be defined relative to the robot.

For environments with parameterized skills and concepts implemented, we can also use planners to derive demonstrations at scale.
As part of \gymname{}, we use a combination of teleoperation and planning to provide $\ge100$ precollected demonstrations for 10 environments (Appendix~\ref{app:kindergym-additional-details}).
We encourage \projectname{} users to collect and open-source additional demonstrations using the teleoperation interfaces provided.

\section{\benchmarkname: Baselines and Metrics}
\label{sec:kinderbench}

Our third contribution is \benchmarkname{}, a standardized multi-metric benchmark for robot physical reasoning.
We report results and release implementations for 13 baselines in 8 environments.
In this section, we briefly describe the environments, baselines, and metrics, analyze the results, and present insights from additional experiments.

\paragraph{Environments}
We select two representative environments with varying levels of difficulty from each of the four \gardenname{} categories.
See Appendix~\ref{app:environments} for detailed descriptions.

\begin{enumerate}
    \item $\mathrm{Motion2D}$: The simplest Kinematic2D environment. The robot must move to reach a goal region. In this variant ($\mathrm{p0}$), there are no obstacles.
    \item $\mathrm{StickButton2D}$: A Kinematic2D environment where a robot must press a button. The button is sometimes out of reach, requiring the robot to use a stick as a tool to press it. In this variant ($\mathrm{b1}$), there is one button.
    \item $\mathrm{DynObstruction2D}$: The Dynamic2D environment in Figure~\ref{fig:kin-vs-dyn-reasoning}. In this variant ($\mathrm{o1}$), there is one obstacle.
    \item $\mathrm{DynPushPullHook2D}$: A Dynamic2D environment that requires using a hook to pull a target object surrounded by obstacles. In this variant ($\mathrm{o5}$), there are five obstacles.
    \item $\mathrm{BaseMotion3D}$: The simplest Kinematic3D environment. The robot must move its base to reach a goal region. In this variant ($\mathrm{o0}$), there are no obstacles.
    \item $\mathrm{Transport3D}$: A Kinematic3D environment where a box and one or more objects must be moved from the floor to a table. The box may be used as a container, but this is not required (and not always optimal). In this variant ($\mathrm{o2}$), there are two objects in addition to the box.
    \item $\mathrm{Shelf3D}$: A Dynamic3D environment where objects must be packed into a space-constrained shelf. In this variant ($\mathrm{o1}$), one object must be packed.
    \item $\mathrm{SweepIntoDrawer3D}$: A Dynamic3D environment where small objects on a countertop must be moved to an initially closed drawer, optionally using a sweeping tool. In this variant ($\mathrm{o5}$), there are 5 objects.
\end{enumerate}

\paragraph{Baselines}

We evaluate 13 representative planning and learning baselines.
See Appendix~\ref{app:bench-additional-details} for details.

\begin{enumerate}
    \item \textbf{Bilevel Planning~(BP)}~\cite{srivastava2014combined,silver2023predicateinvent,li2025IVNTR}: A search-then-sample TAMP planner; uses skills and concepts.
    \item \textbf{LLM Planning (LLMPlan)}~\cite{silver2022pddl,song2023llm}: An LLM (GPT-5.2~\cite{singh2025openai}) planner; uses object-centric states and skills.
    \item \textbf{VLM Planning (VLMPlan)}~\cite{hu2023look}: A VLM (GPT-5.2~\cite{singh2025openai}) planner; uses \emph{RGB images}, states, and skills.
    \item \textbf{LLM In-context (LLMCon)}~\cite{silver2022pddl,song2023llm}: An LLM (GPT-5.2~\cite{singh2025openai}) planner that is prompted with in-context examples and uses object-centric states and skills.
    \item \textbf{VLM In-context (VLMCon)}~\cite{hu2023look}: A VLM (GPT-5.2~\cite{singh2025openai}) planner that is prompted with in-context examples; uses \emph{RGB images} along with states and skills.
    \item \textbf{Model Predictive Control~(MPC)}~\cite{howell2022predictive}: We perform predictive sampling trajectory optimization using MPC with the ground-truth simulation transition functions and (sparse) reward function. 
    \item \textbf{Model-based Reinforcement Learning~(MBRL)}~\cite{chitnis2022learning}: We use the demonstrations collected to train a neural (state-based) transition model and use MPC to select actions with the same sparse reward function.
    \item \textbf{Generative Diffusion Planning}: We train and evaluate Generative Skill Chaining (GSC)~\cite{mishra2023gsc} with our demonstration data annotated with skill labels.
    \item \textbf{Proximal Policy Optimization (PPO)}~\cite{schulman2017proximal}: A standard \emph{on-policy} deep reinforcement learning method.
    \item \textbf{Soft Actor-Critic (SAC)}~\cite{haarnoja2018soft}: A standard \emph{off-policy} deep reinforcement learning method.
    \item \textbf{Diffusion Policy (DP)}~\cite{chi2025diffusion}: Imitation learning over RGB images, trained with 100 demos per environment.
    \item \textbf{DP + Environment States (DPES)}~\cite{chi2025diffusion}: Same as DP, but with states also provided as input.
    \item \textbf{Finetuned VLA}~\cite{intelligence2025pi_}: A pretrained $\pi_{0.5}$ VLA fine-tuned on the same demos as DP.
\end{enumerate}

\begin{table*}[ht]
\centering
\footnotesize
\setlength{\tabcolsep}{3pt}

\begin{tabular}{lccccccccccccc}
\toprule[1.5pt]
Method & BP & LLMCon & VLMCon & LLMPlan & VLMPlan & MPC & VLA & GSC & DPES & DP & PPO & MBRL & SAC\\
\midrule

\multicolumn{14}{c}{\textbf{Motion2D (Kinematic2D)}}\\
SR $\uparrow$        & \textbf{1.00} & \textbf{1.00} & \textbf{1.00} & 0.99 & \textbf{1.00} & 0.92 & 0.79 & \textbf{1.00} & 0.92 & 0.88 & 0.80 & 0.16 & 0.00 \\
Rwd $\uparrow$       & -40.0 & -40.0 & -40.0 & -39.9 & -39.9 & -92.0 & -41.9 & -54.9 & -45.0 & -44.8 & -39.6 & -186.4 & -- \\
Inf-Time $\downarrow$& 0.07 & 0.02 & 2.62 & 2.72 & 2.89 & 29.20 & 0.62 & 13.20 & 0.23 & 0.28 & 0.002 & 72.10 & -- \\

\midrule
\multicolumn{14}{c}{\textbf{StickButton2D (Kinematic2D)}}\\
SR $\uparrow$        & \textbf{0.99} & 0.44 & 0.44 & 0.25 & 0.28 & 0.68 & 0.53 & 0.08 & 0.10 & 0.10 & 0.14 & 0.36 & 0.00 \\
Rwd $\uparrow$       & -52.6 & -43.3 & -46.1 & -62.9 & -63.2 & -90.3 & -52.0 & -383.2 & -48.6 & -72.4 & -17.0 & -149.2 & -- \\
Inf-Time $\downarrow$& 0.85 & 2.14 & 2.93 & 2.49 & 3.09 & 117.60 & 0.97 & 161.70 & 0.66 & 0.67 & 0.01 & 280.80 & -- \\

\midrule
\multicolumn{14}{c}{\textbf{DynObstruction2D (Dynamic2D)}}\\
SR $\uparrow$        & 0.08 & 0.07 & 0.07 & 0.02 & 0.03 & 0.41 & \textbf{0.50} & 0.32 & 0.34 & 0.33 & 0.08 & 0.04 & 0.01 \\
Rwd $\uparrow$       & -68.62 & -82.6 & -84.1 & -30.2 & -5.5 & -142.8 & -82.7 & -408.9 & -105.1 & -96.4 & -29.2 & -195.8 & -1.0 \\
Inf-Time $\downarrow$& 15.95 & 5.39 & 3.83 & 3.92 & 4.19 & 381.00 & 1.13 & 202.50 & 0.58 & 0.58 & 0.01 & 571.92 & 0.02 \\

\midrule
\multicolumn{14}{c}{\textbf{DynPushPullHook2D (Dynamic2D)}}\\
SR $\uparrow$        & 0.01 & 0.01 & 0.01 & 0.00 & 0.00 & 0.00 & \textbf{0.43} & 0.00 & 0.00 & 0.00 & 0.00 & 0.00 & 0.00 \\
Rwd $\uparrow$       & -105.3 & -117.0 & -94.3 & -- & -- & -- & -253.8 & -- & -- & -- & -- & -- & -- \\
Inf-Time $\downarrow$& 73.8 & 5.58 & 7.81 & -- & -- & -- & 3.23 & -- & -- & -- & -- & -- & -- \\

\midrule
\multicolumn{14}{c}{\textbf{BaseMotion3D (Kinematic3D)}}\\
SR $\uparrow$        & \textbf{1.00} & \textbf{1.00} & \textbf{1.00} & \textbf{1.00} & \textbf{1.00} & 0.54 & 0.25 & 0.66 & 0.54 & 0.32 & 0.00 & 0.06 & 0.15 \\
Rwd $\uparrow$       & -9.9 & -9.9 & -9.9 & -9.9 & -9.9 & -70.1 & -26.5 & -93.0 & -28.7 & -26.8 & -- & -95.7 & -7.8 \\
Inf-Time $\downarrow$& 0.02 & 2.54 & 0.01 & 2.04 & 2.16 & 58.00 & 0.66 & 164.80 & 0.28 & 0.35 & -- & 125.54 & 0.003 \\

\midrule
\multicolumn{14}{c}{\textbf{Transport3D (Kinematic3D)}}\\
SR $\uparrow$        & \textbf{0.46} & 0.36 & 0.34 & 0.43 & 0.38 & 0.00 & 0.00 & 0.00 & 0.00 & 0.00 & 0.00 & 0.00 & 0.00 \\
Rwd $\uparrow$       & -899.0 & -626.2 & -607.4 & -695.3 & -657.5 & -- & -- & -- & -- & -- & -- & -- & -- \\
Inf-Time $\downarrow$& 21.4 & 3.96 & 4.66 & 3.06 & 1.68 & -- & -- & -- & -- & -- & -- & -- & -- \\

\midrule
\multicolumn{14}{c}{\textbf{Shelf3D (Dynamic3D)}}\\
SR $\uparrow$        & \textbf{1.00} & 0.55 & 0.52 & 0.00 & 0.00 & 0.00 & 0.02 & 0.00 & 0.09 & 0.13 & 0.00 & 0.00 & 0.00 \\
Rwd $\uparrow$       & -99.9 & -105.9 & -110.0 & -- & -- & -- & -342.7 & -- & -356.0 & -364.7 & -- & -- & -- \\
Inf-Time $\downarrow$& 1.59 & 3.45 & 3.68 & -- & -- & -- & 3.90 & -- & 2.63 & 2.52 & -- & -- & -- \\

\midrule
\multicolumn{14}{c}{\textbf{SweepIntoDrawer3D (Dynamic3D)}}\\
SR $\uparrow$        & 0.03 & 0.00 & 0.00 & 0.00 & 0.00 & 0.00 & 0.00 & 0.00 & 0.04 & \textbf{0.14} & 0.00 & 0.00 & 0.00 \\
Rwd $\uparrow$       & -157.4 & -- & -- & -- & -- & -- & -- & -- & -516.6 & -520.6 & -- & -- & -- \\
Inf-Time $\downarrow$& 28.3 & -- & -- & -- & -- & -- & -- & -- & 3.59 & 3.13 & -- & -- & -- \\

\bottomrule[1.5pt]
\end{tabular}
\caption{Benchmark evaluations across representative Kinematic2D, Dynamic2D, Kinematic3D, and Dynamic3D environments.}
\label{tab:emp_all}
\vspace{-5pt}
\end{table*}

\paragraph{Evaluation Metrics}

\benchmarkname{} includes multiple metrics that capture different dimensions of efficiency and effectiveness in physical reasoning.
These include:
\begin{enumerate}
    \item \textbf{Success Rate~(SR)}: A measure of effectiveness.
    \item \textbf{Cumulative Rewards~(Rwd)}: A measure of efficiency. Recall rewards are $-1$ until success. This metric is considered only for successful episodes.
    \item \textbf{Inference Time~(Inf-Time)}: Another measure of efficiency. We report per-episode wall-clock time (sec).
\end{enumerate}

Another metric that is equally important but difficult to capture quantitatively is \textbf{engineering cost}: the amount of engineering needed to run a method in a new environment.
For example, BP has a high engineering cost because it requires skills and concepts; RL has a much lower cost.

\paragraph{Benchmark Results}
We present our main benchmark results in Table~\ref{tab:emp_all}.
All baselines are evaluated over 5 random seeds with 50 evaluation episodes per seed. 
We report means in the main paper and standard deviations in Appendix~\ref{app:bench-additional-details}.
Overall, BP obtains the highest average success rate (0.57), followed by LLMCon~(0.43), VLMCon~(0.43), LLMPlan~(0.34), VLMPlan~(0.34), MPC~(0.32), VLA~(0.32), GSC~(0.26), DPES~(0.25), DP~(0.24), PPO~(0.13), MBRL~(0.08), SAC~(0.02). 
The general trend is expected: paying higher engineering costs and spending more inference time leads to dividends in success rates.

We now discuss baseline performance in detail, highlighting some of the more surprising results.
First, given that both use the same parameterized skills, the gap between BP and LLMPlan/VLMPlan, especially in the more challenging environments, indicates that there remains room to improve the latter approaches.
The comparison between LLMPlan/VLMPlan and LLMCon/VLMCon shows that in-context examples are important for the overall performance. 
Furthermore, the comparable performance between LLMPlan/VLMPlan suggests that the VLM is not able to meaningfully leverage the images that it receives in addition to the object-centric states.

\begin{table}[!t]
    \centering
    \setlength{\tabcolsep}{0.6mm}
    \fontsize{7.5}{9}\selectfont
    \begin{tabular}{c|cccc}
    \toprule[1.5pt]
    Subtask & Open Drawer & Grasp Sweeper & Sweep Some Objects & Sweep All Objects \\
    \midrule
    DPES   & 0.50 & 0.28  & 0.08  & 0.04 \\
    DP    & 0.87 & 0.74  & 0.22  & 0.14 \\
    VLA   & 0.01     & 0.00     & 0.00     & 0.00 \\
    \bottomrule[1.5pt]
    \end{tabular}%
  \caption{Subtask success rate for DPES, DP, VLA baselines in the $\mathrm{SweepIntoDrawer3D}$ environment.}
  \label{tab:subtask}%
  \vspace{-10pt}
\end{table}%

The imitation learning baselines (DP, DPES, VLA) perform well overall, considering that they do not have access to parameterized skills.
Interestingly, the VLA is the only baseline that achieves a non-trivial success rate (0.43) on the $\mathrm{DynPushPullHook2D}$ task with 5 obstacles.
Recall that this environment requires both tool use and nonprehensile multi-object manipulation. 
This result is surprising because the 2D rendering and physics is quite different from the data used to pretrain the VLA.
Another surprising finding is the nontrivial success rate of DP (0.14) and DPES (0.04) on the long-horizon multi-stage $\mathrm{SweepIntoDrawer3D}$, which requires the robot to open the drawer, grasp the sweeper, and then sweep multiple objects into the drawer. 
We also provide subtask success rates in Table~\ref{tab:subtask}. 
We also see that DPES performs comparably to DP, despite its access to object-centric states.
This suggests that DP is not able to meaningfully leverage the states, which is an interesting dual to the LLM/VLM case.

We also find that the MPC baseline performs well, given that it only receives the sparse reward functions. 
We attribute the good performance to the predictive sampling proposed in~\cite{howell2022predictive}. 
The MBRL performs worse than the MPC baseline, though they use the same planner, which demonstrates that the learned transition model is unreliable.

Finally, we find that the RL baselines (PPO and SAC) perform well only in short-horizon tasks, which is expected given the sparse rewards and lack of inductive bias given to these methods.
In Appendix~\ref{app:kindergym-additional-details}, we report additional results showing that dense rewards can improve performance, but success rates for RL remain low overall.

\begin{table}[ht]
\centering
\setlength{\tabcolsep}{0.6mm}
    \fontsize{9.5}{12}\selectfont
    \begin{tabular}{ccccccccc}
    \toprule[1.5pt]
    Baselines & \multicolumn{4}{c}{DP} & \multicolumn{4}{c}{VLA} \\
    \midrule
    Num. Obstacles & 1     & 0     & 2  & 3   & 1     & 0     & 2  & 3\\
    SR $\uparrow$ & 0.33 & 0.35 & 0.30 & 0.26 & 0.50 & 0.52 & 0.43 & 0.40\\
    \bottomrule[1.5pt]
    \end{tabular}%
    \caption{$\mathrm{DynObstruction2D}$ out-of-distribution generalization. Training has 1 obstacle; evaluation has 0-3 obstacles.}
    \label{tab:bc_objects}
\end{table}%

\paragraph{Additional Results}

\textbf{Out-of-Distribution Generalization:}
We next evaluate generalization to unseen scenarios for the imitation learning baselines (DP, VLA) that do not require environment-specific state vectors in the $\mathrm{DynObstruction2D}$ environment.
After training with 1 obstacle, we test with 0, 2, and 3 obstacles.
Results are shown in Table~\ref{tab:bc_objects}.
Although there is some performance degradation, both baselines perform surprisingly well in these out-of-distribution tasks.
The VLA is particularly robust, perhaps due to pretraining.

\begin{table}[ht]
\centering
\setlength{\tabcolsep}{1.6mm}
    \fontsize{8}{12}\selectfont
    \begin{tabular}{ccccc}
    \toprule[1.5pt]
    Num. Buttons & 1     & 3     & 5     & 10 \\
    \midrule
    SR $\uparrow$ & 0.99  & 0.26  & 0.02  & 0.00 \\
    Rwd $\uparrow$ & -52.60 & -86.30 & -123.80 & -- \\
    Inf-Time $\downarrow$ & 1.51  & 19.45 & 28.12 & 38.39 \\
    \bottomrule[1.5pt]
    \end{tabular}%
    \caption{Test-time generalization for bilevel planning baseline in the $\mathrm{StickButton2D}$ environment.}
    \label{tab:planning_objects}
    \vspace{-1.6em}
\end{table}%

\textbf{Scaling Bilevel Planning:}
We finally evaluate the efficiency and effectiveness of bilevel planning in the $\mathrm{StickButton2D}$ environment as the number of objects increases.
In Table~\ref{tab:planning_objects}, we see that the success rate and planning time substantially decrease and increase respectively.
This highlights an opportunity for future work that uses learning to improve planning for physical reasoning at scale.

\section{Real World Validation}
\label{sec:real-world}
\begin{figure}[ht]
    \centering
    \vspace{-1em}\includegraphics[width=\linewidth]{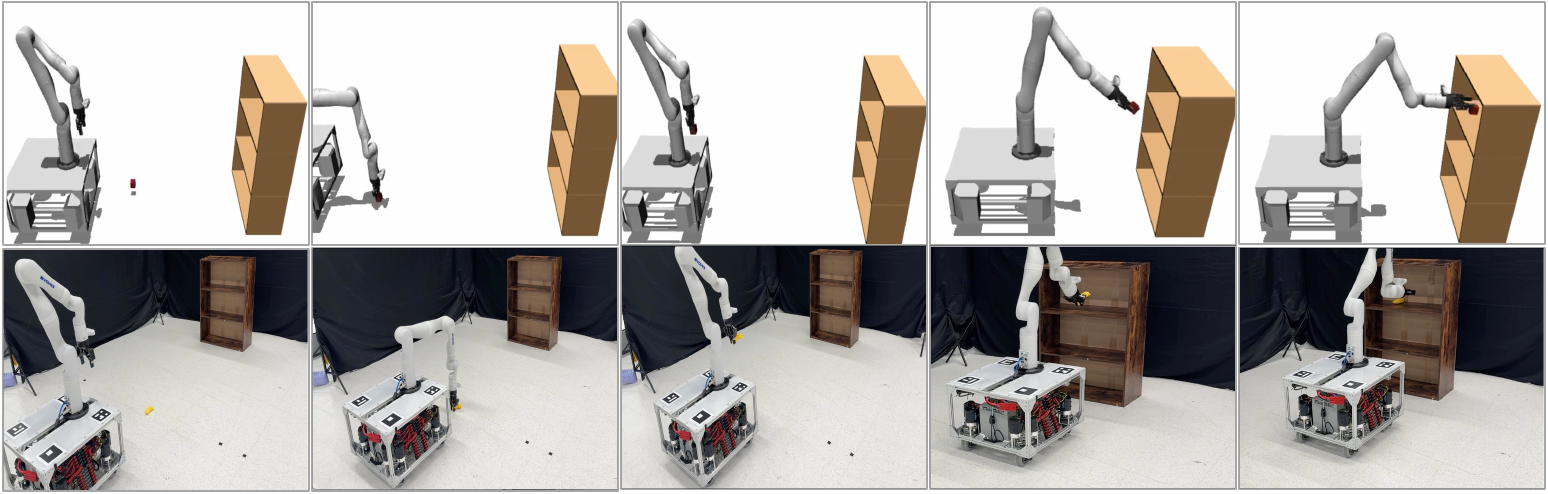}
    \caption{Real-to-sim-to-real example. We construct a twin simulation from real-world observations using object-centric states, generate motion plans in simulation, and execute them in the real world.}
    \label{fig:rs2r2}
    \vspace{-10pt}
\end{figure}

We next demonstrate an example of real-to-sim-to-real with the TidyBot++ as the real robot~\cite{wu2024tidybot++} and the $\mathrm{Shelf3D}$ environment in \gardenname{} as the simulator (Figure~\ref{fig:rs2r2}).
Our goal is to show that \gardenname{} corresponds to real-world physical reasoning challenges, while also highlighting the potential for future real-to-sim-to-real research using \projectname{}.
We use an overhead camera to localize the robot and obtain object bounding boxes and poses using~\cite{gu2021open}.
Given the estimated robot and object poses, we initialize the robot states and object-centric states accordingly. 
We then generate a plan in the simulator and execute it back in the real world.
See Appendix~\ref{app:real-to-sim-to-real} for additional discussion.

\section{Limitations and Discussion}
\label{sec:limitations-and-discussion}
We conclude by acknowledging several limitations of the present work. First, as with any simulation-based benchmark, certain aspects of real-world physics and interaction are not fully captured. While the challenges emphasized here primarily target ``mid-level'' reasoning, where fine-grained physical details may be less critical, there exist important dimensions of physical reasoning for which real-world fidelity plays a more substantial role.
Second, to maintain a manageable scope, we made a number of design choices that necessarily exclude other factors relevant to robotics and physical reasoning, including stochasticity, partial observability, diverse robot embodiments, and multi-robot coordination.
Third, although we selected baseline methods that are standard and broadly representative, many alternative approaches were not evaluated. We look forward to actively supporting the community and maintaining \projectname{} open-source as researchers develop and evaluate more sophisticated methods.

\section*{Acknowledgement}
We acknowledge the support of the Air Force Research Laboratory (AFRL), DARPA, under agreement number FA8750-23-2-1015.
We also acknowledge the Defense Science and Technology Agency (DSTA) under contract \#DST000EC124000205.
The authors would also like to express sincere gratitude to Dr. Caelan Garrett (NVIDIA), Dr. Shuangyu Xie (UC Berkeley), Dr. Kaiyuan Chen (UC Berkeley), Prof. David Held (CMU), Prof. Zak Kingston (Purdue University), Dr. Ajay Mandlekar (NVIDIA), Prof. Ben Eysenbach (Princeton), Prof. Jeannette Bohg (Stanford), Prof. Rika Antonova (University of Cambridge), Carlota Parés-Morlans (Stanford), Yishu Li (CMU), Prof. Florian Shkurti (University of Toronto), Prof. Greg Stein (George Mason University), and Prof. George Konidaris (Brown University) for their insightful suggestions and comments when this project was at an early stage.
We thank Google’s TPU Research Cloud (TRC) for providing access to Cloud TPUs.

\bibliographystyle{IEEEtranN}
\bibliography{references}

\clearpage

\appendix

\begin{table*}[ht] \centering \footnotesize \setlength{\tabcolsep}{3pt}  
\begin{tabular}{lccccccccccccc} \toprule[1.5pt] 
Method & BP & LLMCon & VLMCon & LLMPlan & VLMPlan & MPC & VLA & GSC & DPES & DP & PPO & MBRL & SAC \\ \midrule  
\multicolumn{14}{c}{Motion2D (Kinematic2D)} \\ 
SR    & 0.000 & 0.000 & 0.000 & 0.090 & 0.000 & 0.050 & 0.037 & 0.000 & 0.038 & 0.044 & 0.450 & 0.020 & 0.000 \\ 
Rwd   & 0.582 & 0.470 & 0.470 & 0.620 & 0.540 & 7.300 & 1.008 & 13.180 & 0.488 & 0.320 & 0.100 & 2.800 & -- \\ 
Inf-Time & 0.010 & 0.003 & 0.460 & 0.760 & 0.400 & 2.100 & 0.109 & 6.050 & 0.001 & 0.001 & 0.001 & 1.660 & -- \\  
\midrule \multicolumn{14}{c}{StickButton2D (Kinematic2D)} \\ 
SR    & 0.089 & 0.500 & 0.500 & 0.430 & 0.450 & 0.040 & 0.0676 & 0.278 & 0.013 & 0.040 & 0.150 & 0.040 & 0.000 \\ 
Rwd   & 39.826 & 33.600 & 33.900 & 38.700 & 38.900 & 8.000 & 8.250 & 26.350 & 15.700 & 14.340 & 7.390 & 8.400 & -- \\ 
Inf-Time & 4.129 & 0.490 & 0.680 & 0.430 & 0.520 & 10.630 & 0.036 & 89.940 & 0.006 & 0.005 & 0.000 & 15.300 & -- \\  
\midrule \multicolumn{14}{c}{DynObstruction2D (Dynamic2D)} \\ 
SR    & 0.270 & 0.260 & 0.260 & 0.150 & 0.170 & 0.060 & 0.034 & 0.466 & 0.053 & 0.097 & 0.045 & 0.02 & 0.007 \\ 
Rwd   & 22.69 & 47.800 & 46.700 & 54.200 & 19.800 & 13.790 & 2.800 & 91.110 & 9.400 & 6.100 & 8.590 & 3.960 & 0.000 \\ 
Inf-Time & 21.96 & 3.070 & 2.700 & 0.830 & 0.670 & 33.950 & 0.107 & 52.210 & 0.008 & 0.008 & 0.000 & 12.010 & 0.000 \\  
\midrule \multicolumn{14}{c}{DynPushPullHook2D (Dynamic2D)} \\ 
SR    & 0.278 & 0.110 & 0.100 & 0.000 & 0.000 & 0.000 & 0.053 & 0.000 & 0.000 & 0.000 & 0.000 & 0.000 & 0.000 \\ 
Rwd   & 23.010 & 47.400 & 26.800 & -- & -- & -- & 8.630 & -- & -- & -- & -- & -- & -- \\ 
Inf-Time & 20.869 & 3.230 & 22.300 & -- & -- & -- & 0.067 & -- & -- & -- & -- & -- & -- \\  
\midrule \multicolumn{14}{c}{BaseMotion3D (Kinematic3D)} \\ 
SR    & 0.000 & 0.000 & 0.000 & 0.000 & 0.000 & 0.040 & 0.040 & 0.473 & 0.040 & 0.080 & 0.007 & 0.020 & 0.132 \\ 
Rwd   & 0.366 & 0.550 & 0.550 & 0.550 & 0.550 & 3.820 & 1.240 & 31.190 & 1.860 & 1.880 & 0.000 & 1.800 & 3.980 \\ 
Inf-Time & 0.004 & 0.560 & 0.001 & 0.420 & 0.830 & 3.060 & 0.050 & 125.510 & 0.003 & 0.003 & 0.000 & 2.600 & 0.004 \\  
\midrule \multicolumn{14}{c}{Transport3D (Kinematic3D)} \\ 
SR    & 0.499 & 0.480 & 0.470 & 0.500 & 0.490 & 0.000 & 0.000 & 0.000 & 0.000 & 0.000 & 0.000 & 0.000 & 0.000 \\ 
Rwd   & 39.826 & 496.000 & 522.000 & 465.000 & 487.000 & -- & -- & -- & -- & -- & -- & -- & -- \\ 
Inf-Time & 4.764 & 1.600 & 1.900 & 2.000 & 3.000 & -- & -- & -- & -- & -- & -- & -- & -- \\  
\midrule \multicolumn{14}{c}{Shelf3D (Dynamic3D)} \\ 
SR    & 0.000 & 0.500 & 0.500 & 0.000 & 0.000 & 0.000 & 0.015 & 0.000 & 0.027 & 0.063 & 0.000 & 0.000 & 0.000 \\ 
Rwd   & 4.900 & 0.110 & 0.140 & -- & -- & -- & 0.158 & -- & 0.154 & 0.212 & -- & -- & -- \\ 
Inf-Time & 0.700 & 1.780 & 0.720 & -- & -- & -- & 0.068 & -- & 0.007 & 0.041 & -- & -- & -- \\  
\midrule \multicolumn{14}{c}{SweepIntoDrawer3D (Dynamic3D)} \\ 
SR    & 0.18 & 0.000 & 0.000 & 0.000 & 0.000 & 0.000 & 0.000 & 0.000 & 0.026 & 0.037 & 0.000 & 0.000 & 0.000 \\ 
Rwd   & 4.900 & -- & -- & -- & -- & -- & -- & -- & 41.000 & 25.700 & -- & -- & -- \\ 
Inf-Time & 19.500 & -- & -- & -- & -- & -- & -- & -- & 0.087 & 0.007 & -- & -- & -- \\  
\bottomrule[1.5pt] 
\end{tabular} 
\caption{Standard deviations of the benchmark evaluations across representative Kinematic2D, Dynamic2D, Kinematic3D, and Dynamic3D environments.} 
\label{tab:emp_std} 
\vspace{-2pt} 
\end{table*}

\subsection{\gardenname{} Environment Details}
\label{app:environments}

\begin{table}[t]
\centering
\setlength{\tabcolsep}{1.6mm}
    \fontsize{9.5}{12}\selectfont
    \begin{tabular}{c|ccc|ccc}
    \toprule[1.5pt]
    Algo.  & \multicolumn{3}{c|}{PPO} & \multicolumn{3}{c}{SAC} \\
    Metric & Dense & Sparse & Delta & Dense & Sparse & Delta \\
    \midrule
    SR  $\uparrow$  & 0.10  & 0.00  & -1.00  & 0.10  & 0.15  & +0.67 \\
    Rwd $\uparrow$ & -28.69  & N/A  & -1.00  & -13.89  & -7.78  & +0.78 \\
    \bottomrule[1.5pt]
    \end{tabular}%
    \caption{Comparison between dense reward and sparse reward settings for RL baselines in BaseMotion3D.}
    \label{tab:rl_rwd}
\end{table}%

\begin{table}[ht]
\centering
    \small
    \setlength{\tabcolsep}{1.6mm}
    \fontsize{9.5}{12}\selectfont
    \begin{tabular}{lcccccc}
    \toprule[1.5pt]
    Noise Type & \multicolumn{3}{c}{Observation} & \multicolumn{3}{c}{Action} \\
    \midrule
    Noise (stdev) & 0     & 0.01  & 0.1   & 0     & 0.01  & 0.1 \\
    StickButton2D & 1.00     & 0.22  & 0.06  & 1.00     & 0.41  & 0.05 \\
    Motion2D & 1.00     & 0.40   & 0.08  & 1.00     & 0.13  & 0.00 \\
    \bottomrule[1.5pt]
    \end{tabular}%
    \caption{Bilevel planning with noise wrappers.}
    \label{tab:noise}
    \vspace{-1em}
\end{table}

\begin{table*}[t]
\centering
\caption{Details about the Kinematic 2D Environments.}
\label{tab:kinematic2d_domains}
\renewcommand{\arraystretch}{1.2}

\begin{adjustbox}{width=\textwidth}
\begin{tabular}{
  >{\centering\arraybackslash}m{0.16\textwidth}
  m{0.36\textwidth}
  m{0.24\textwidth}
  m{0.12\textwidth}
  >{\centering\arraybackslash}m{0.12\textwidth}
}
\toprule[1.5pt]
\textbf{Domain} &
\textbf{Description} &
\textbf{Core Challenges} &
\textbf{Variants} &
\textbf{State / Action} \\
\midrule

\includegraphics[width=0.14\textwidth]{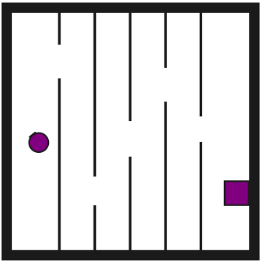}
&
\textbf{Motion2D.}
A 2D environment where the goal is to reach a target region while avoiding static obstacles.
There may be narrow passages.
The robot has a movable circular base and a retractable arm with a rectangular vacuum end effector. 
The arm and vacuum do not need to be used in this environment.
&
\makecell[l]{
\textbf{Kinematic2D} \\
Basic Spatial Relations
}
&
The variants differ in the number of passages.
&
\makecell[c]{
$\mathcal{S} \in \mathbb{R}^{N \times 20 + 19}$ \\
$\mathcal{A} \in \mathbb{R}^{5}$
} \\

\midrule

\includegraphics[width=0.14\textwidth]{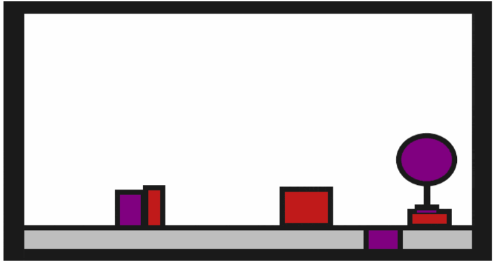}
&
\textbf{Obstruction2D.}
A 2D environment where the goal is to place a target block onto a target surface. The block must be completely contained within the surface boundaries.
The target surface may be initially obstructed.
The robot has a movable circular base and a retractable arm with a rectangular vacuum end effector. Objects can be grasped and ungrasped when the end effector makes contact.
&
\makecell[l]{
\textbf{Kinematic2D} \\
Combinatorial Geometric Constraints
}
&
The variants differ in the number of obstructions.
&
\makecell[c]{
$\mathcal{S} \in \mathbb{R}^{N \times 10 + 29}$ \\
$\mathcal{A} \in \mathbb{R}^{5}$
} \\

\midrule

\includegraphics[width=0.14\textwidth]{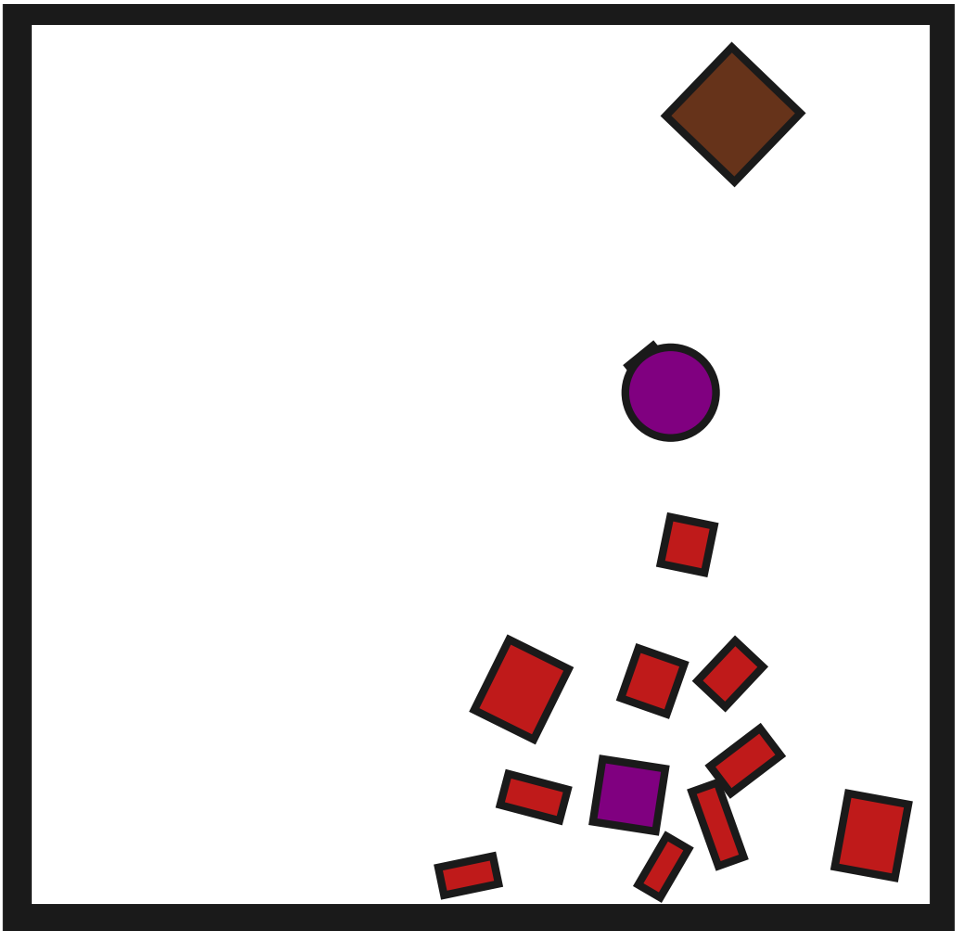}
&
\textbf{ClutteredRetrieval2D.}
A 2D environment where the goal is to retrieve a target block and place it inside a region.
The target block may be initially obstructed by other blocks from all directions.
&
\makecell[l]{
\textbf{Kinematic2D} \\
Combinatorial Geometric Constraints
}
&
The variants differ in the number of obstructions.
&
\makecell[c]{
$\mathcal{S} \in \mathbb{R}^{N \times 10 + 29}$ \\
$\mathcal{A} \in \mathbb{R}^{5}$
} \\

\midrule

\includegraphics[width=0.14\textwidth]{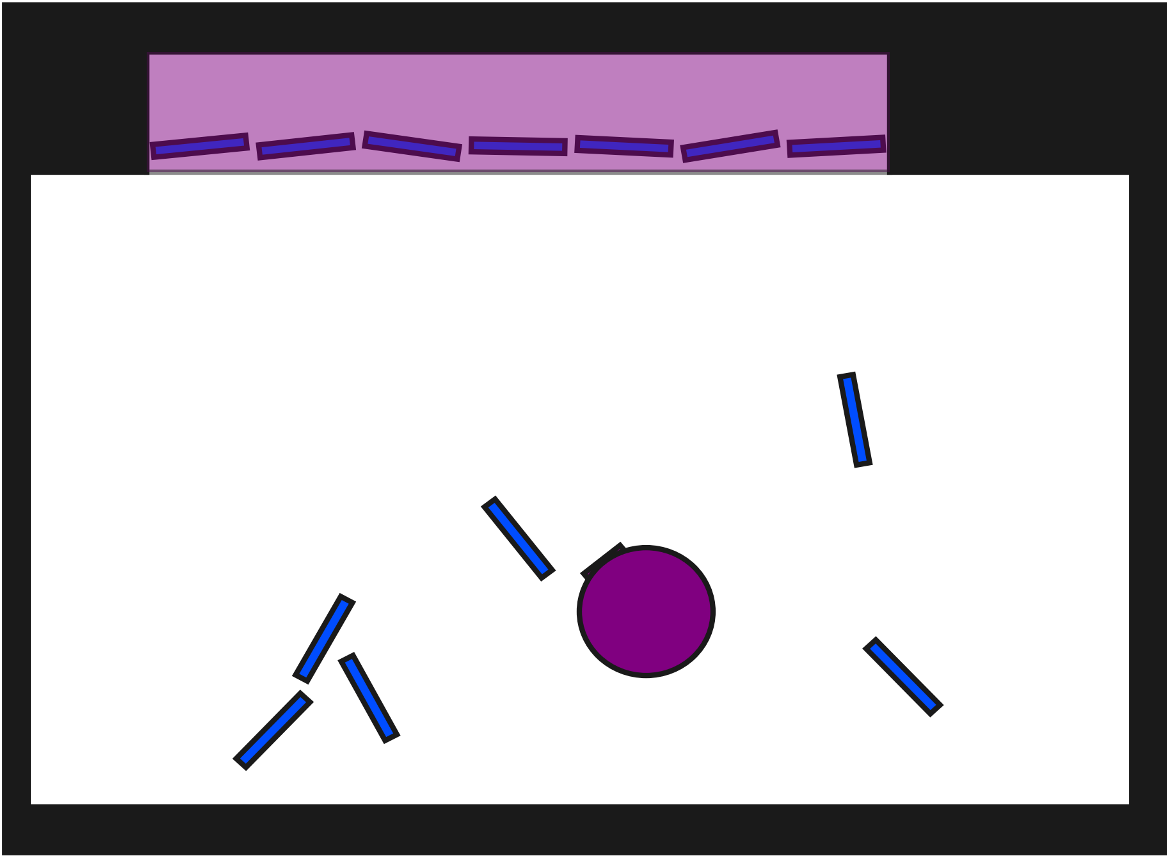}
&
\textbf{ClutteredStorage2D.}
A 2D environment where the goal is to put all blocks inside a shelf.
There may be blocks that are initially inside the shelf, but that need to be rearranged to make space for other blocks.
&
\makecell[l]{
\textbf{Kinematic2D} \\
Combinatorial Geometric Constraints
}
&
The variants differ in the number of blocks.
&
\makecell[c]{
$\mathcal{S} \in \mathbb{R}^{N \times 10 + 28}$ \\
$\mathcal{A} \in \mathbb{R}^{5}$
} \\

\midrule

\includegraphics[width=0.14\textwidth]{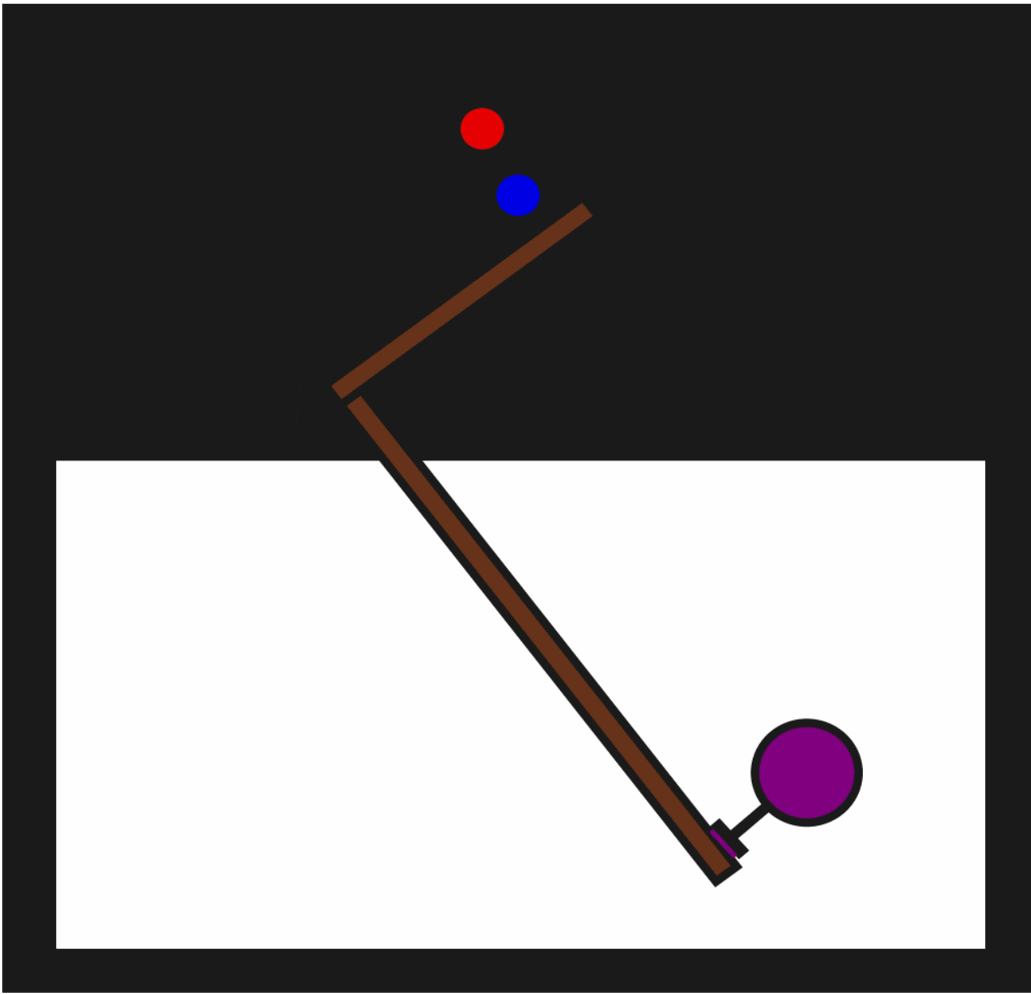}
&
\textbf{PushPullHook2D.}
A 2D environment with a robot, a hook (L-shape), a movable button, and a target button. The robot can use the hook to push the movable button towards the target button. The movable button only moves if the hook is in contact and the robot moves in the direction of contact.
&
\makecell[l]{
\textbf{Kinematic2D} \\
Tool Use
}
&
This environment has one variant.
&
\makecell[c]{
$\mathcal{S} \in \mathbb{R}^{38}$ \\
$\mathcal{A} \in \mathbb{R}^{5}$
} \\

\midrule

\includegraphics[width=0.14\textwidth]{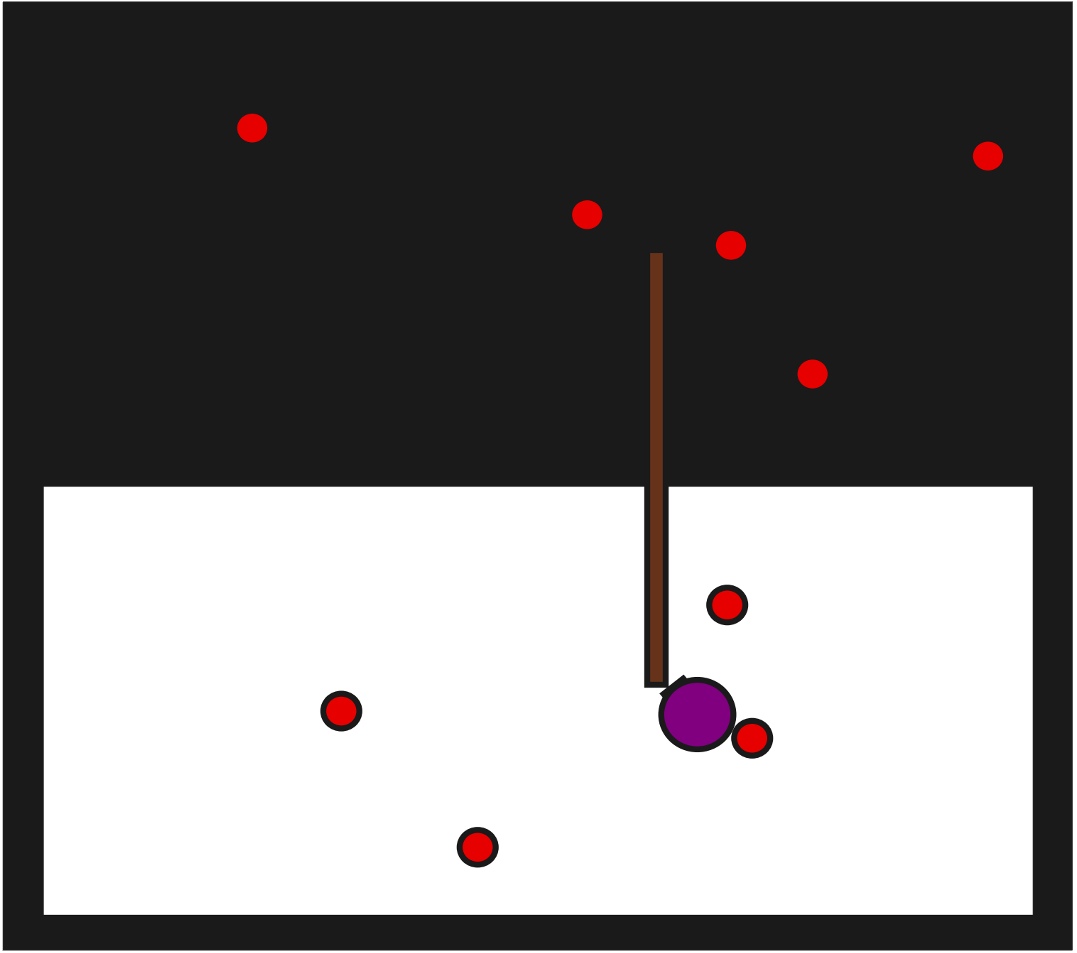}
&
\textbf{StickButton2D.}
A 2D environment where the goal is to touch all buttons, possibly by using a stick for buttons that are out of the robot's direct reach.
This environment is based on the one introduced by~\citet{silver23a}.
&
\makecell[l]{
\textbf{Kinematic2D} \\
Tool Use
}
&
The variants differ in the number of buttons.
&
\makecell[c]{
$\mathcal{S} \in \mathbb{R}^{N\times9+19}$ \\
$\mathcal{A} \in \mathbb{R}^{5}$
} \\

\bottomrule[1.5pt]
\end{tabular}
\end{adjustbox}
\end{table*}

\begin{table*}[t]
\centering
\caption{Details about the Dynamic 2D Environments.}
\label{tab:dynamic2d_domains}
\renewcommand{\arraystretch}{1.2}

\begin{adjustbox}{width=\textwidth}
\begin{tabular}{
  >{\centering\arraybackslash}m{0.16\textwidth}
  m{0.36\textwidth}
  m{0.24\textwidth}
  m{0.12\textwidth}
  >{\centering\arraybackslash}m{0.12\textwidth}
}
\toprule[1.5pt]
\textbf{Domain} &
\textbf{Description} &
\textbf{Core Challenges} &
\textbf{Variants} &
\textbf{State / Action} \\
\midrule

\includegraphics[width=0.14\textwidth]{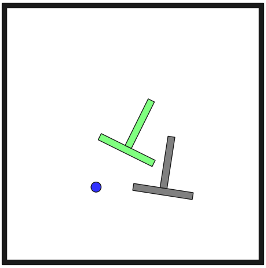}
&
\textbf{DynPushT.}
A 2D physics-based environment where the goal is to push a T-shaped block to match a goal pose using a simple dot robot (kinematic circle) with PyMunk physics simulation.
The T-shaped block must be positioned within small position and orientation thresholds of the goal. This environment is based on the PushT environment from~\cite{chi2025diffusion}.
&
\makecell[l]{
\textbf{Dynamic2D} \\
Basic Spatial Relations
}
&
There is only one variant of this environment.
&
\makecell[c]{
$\mathcal{S} \in \mathbb{R}^{45}$ \\
$\mathcal{A} \in \mathbb{R}^{5}$
} \\

\midrule

\includegraphics[width=0.14\textwidth]{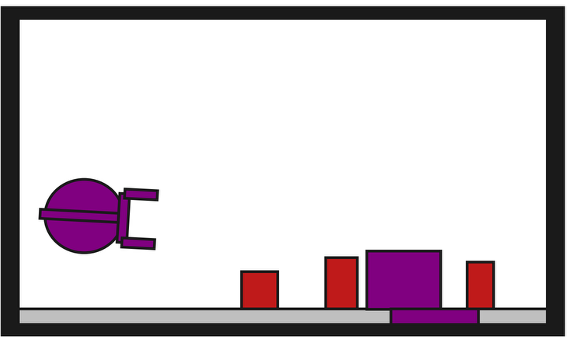}
&
\textbf{DynObstruction2D.}
A 2D physics-based environment where the goal is to place a target block onto a target surface using a two-fingered robot with PyMunk physics simulation. The block must be completely on the surface. The target surface may be initially obstructed.
The robot has a movable circular base and an extendable arm with gripper fingers. Objects can be grasped and released through gripper actions. All objects follow realistic physics, including gravity, friction, and collisions.
&
\makecell[l]{
\textbf{Dynamic2D} \\
Combinatorial Geometric Constraints \\
Nonprehensile Multi-Object Manip.
}
&
The variants differ in the number of obstructions.
&
\makecell[c]{
$\mathcal{S} \in \mathbb{R}^{N \times 15 + 53}$ \\
$\mathcal{A} \in \mathbb{R}^{5}$
} \\

\midrule

\includegraphics[width=0.14\textwidth]{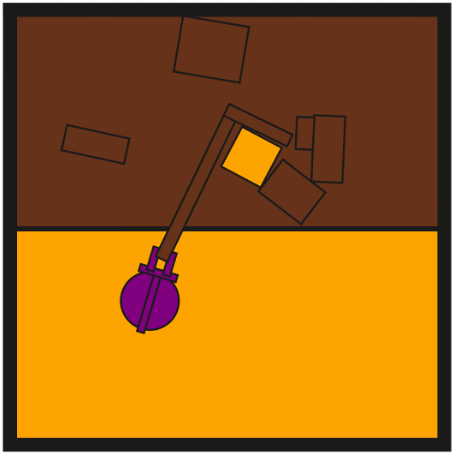}
&
\textbf{DynPushPullHook2D.}
A 2D physics-based tool-use environment where a robot must use a hook to push/pull a target block onto a middle wall (goal surface). The target block is positioned in the upper region of the world, while the middle wall is located at the center. The robot must manipulate the hook to navigate the target block downward through obstacles.
The target block is initially surrounded by obstacle blocks.
The robot has a movable circular base and an extendable arm with gripper fingers. The hook is a kinematic object that can be grasped and used as a tool to indirectly manipulate the target block.
&
\makecell[l]{
\textbf{Dynamic2D} \\
Nonprehensile Multi-Object Manip. \\
Tool Use \\
}
&
The variants differ in the number of obstacles.
&
\makecell[c]{
$\mathcal{S} \in \mathbb{R}^{N \times 14 + 55}$ \\
$\mathcal{A} \in \mathbb{R}^{5}$
} \\

\midrule

\includegraphics[width=0.14\textwidth]{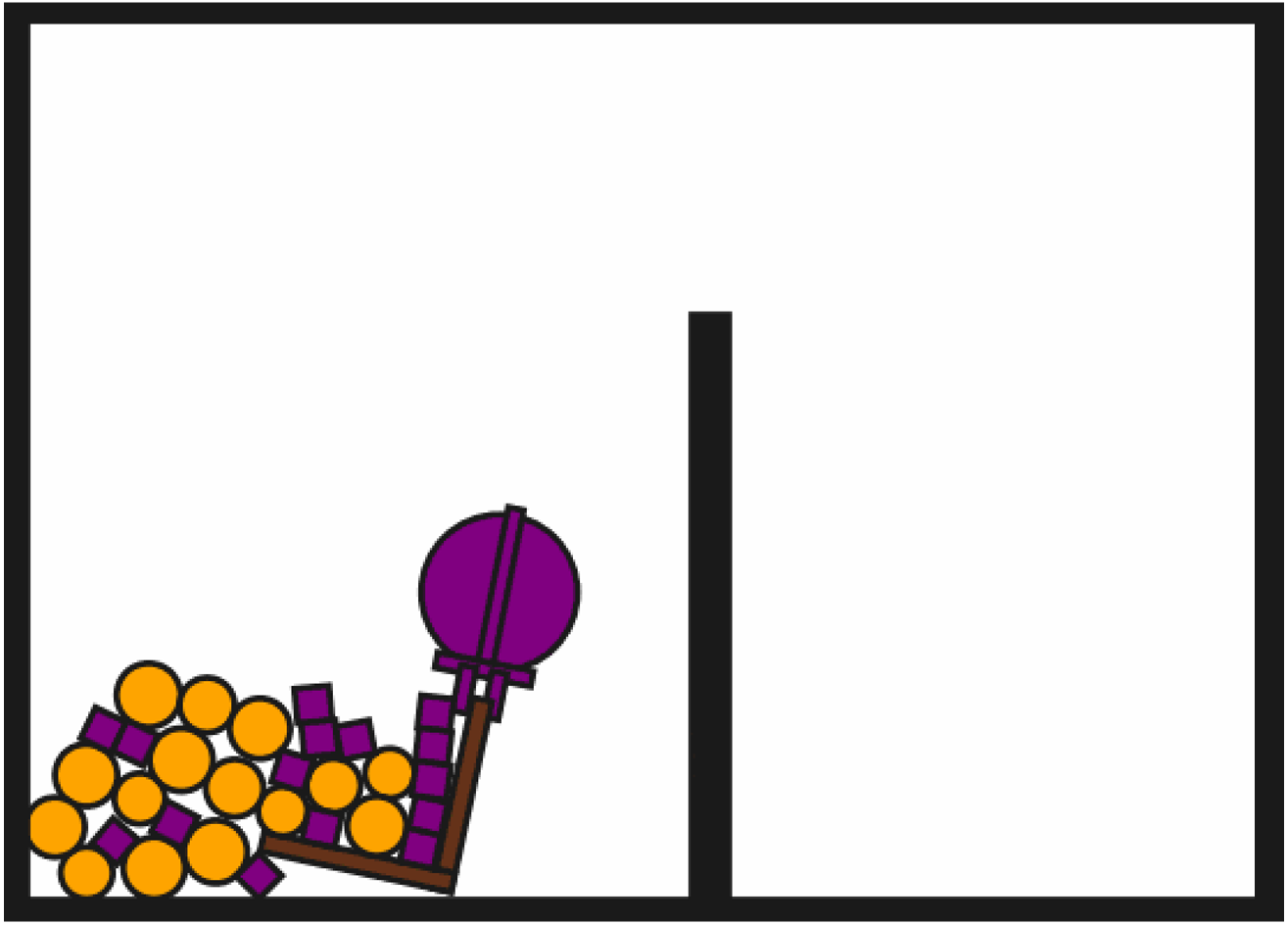}
&
\textbf{ScoopPour2D.}
A 2D physics-based tool-use environment where a robot must use an L-shaped hook to scoop small objects from the left side of a middle wall and pour them onto the right side. The middle wall is half the height of the world, allowing objects to be scooped over it.
The robot has a movable circular base and an extendable arm with gripper fingers. The hook is a kinematic object that can be grasped and used as a tool to scoop the small objects. Small objects are dynamic and follow PyMunk physics, but they cannot be grasped directly by the robot.
&
\makecell[l]{
\textbf{Dynamic2D} \\
Nonprehensile Multi-Object Manip. \\
Tool Use \\
Dynamic Constraints \\
}
&
The variants differ in the number of small objects.
&
\makecell[c]{
$\mathcal{S} \in \mathbb{R}^{N \times 14 + 40}$ \\
$\mathcal{A} \in \mathbb{R}^{5}$
} \\

\bottomrule[1.5pt]
\end{tabular}
\end{adjustbox}
\end{table*}

\begin{table*}[t]
\centering
\caption{Details about the Kinematic 3D Environments.}
\label{tab:kinematic3d_domains}
\renewcommand{\arraystretch}{1.2}

\begin{adjustbox}{width=\textwidth}
\begin{tabular}{
  >{\centering\arraybackslash}m{0.16\textwidth}
  m{0.36\textwidth}
  m{0.24\textwidth}
  m{0.12\textwidth}
  >{\centering\arraybackslash}m{0.12\textwidth}
}
\toprule[1.5pt]
\textbf{Domain} &
\textbf{Description} &
\textbf{Core Challenges} &
\textbf{Variants} &
\textbf{State / Action} \\
\midrule

\includegraphics[width=0.14\textwidth]{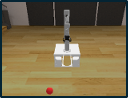}
&
\textbf{BaseMotion3D.}
A mobile manipulator must move its base to reach a goal, with no obstructions.
&
\makecell[l]{
\textbf{Kinematic3D} \\
Basic Spatial Relations
}
&
There is only one variant of this environment.
&
\makecell[c]{
$\mathcal{S} \in \mathbb{R}^{22}$ \\
$\mathcal{A} \in \mathbb{R}^{11}$
} \\

\midrule

\includegraphics[width=0.14\textwidth]{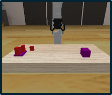}
&
\textbf{Obstruction3D.}
A 3D obstruction clearance environment where the goal is to place a target block on a designated target region by first clearing obstructions.
The robot is a Kinova Gen-3 with 7 degrees of freedom that can grasp and manipulate objects.
&
\makecell[l]{
\textbf{Kinematic3D} \\
Combinatorial Geometric Constraints
}
&
The variants differ in the number of obstructions.
&
\makecell[c]{
$\mathcal{S} \in \mathbb{R}^{N \times 12 + 43}$ \\
$\mathcal{A} \in \mathbb{R}^{11}$
} \\

\midrule

\includegraphics[width=0.14\textwidth]{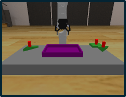}
&
\textbf{Packing3D.}
A 3D packing environment where the goal is to place a set of parts into a rack without collisions.
The parts may be rectangles or triangles.
&
\makecell[l]{
\textbf{Kinematic3D} \\
Combinatorial Geometric Constraints
}
&
The variants differ in the number of parts to pack.
&
\makecell[c]{
$\mathcal{S} \in \mathbb{R}^{N \times 12 + 31}$ \\
$\mathcal{A} \in \mathbb{R}^{11}$
} \\

\midrule

\includegraphics[width=0.14\textwidth]{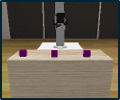}
&
\textbf{Table3D.}
A 3D environment where the goal is to pick up one cube among several that are on a table.
&
\makecell[l]{
\textbf{Kinematic3D} \\
Basic Spatial Relations
}
&
The variants differ in the number of blocks on the table.
&
\makecell[c]{
$\mathcal{S} \in \mathbb{R}^{N \times 12 + 31}$ \\
$\mathcal{A} \in \mathbb{R}^{11}$
} \\

\midrule

\includegraphics[width=0.14\textwidth]{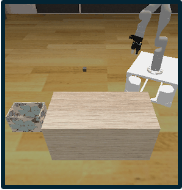}
&
\textbf{Transport3D.}
A 3D environment where the goal is to place all objects, including one or more solid cubes and a box, on a table.
The box may be used as a container, but this is not required, nor is it necessarily optimal.
&
\makecell[l]{
\textbf{Kinematic3D} \\
Tool Use \\
Combinatorial Geometric Constraints
}
&
The variants differ in the number of objects on the ground.
&
\makecell[c]{
$\mathcal{S} \in \mathbb{R}^{N \times 12 + 43}$ \\
$\mathcal{A} \in \mathbb{R}^{11}$
} \\

\bottomrule[1.5pt]
\end{tabular}
\end{adjustbox}
\end{table*}

\begin{table*}[t]
\centering
\caption{Details about the Dynamic 3D Environments.}
\label{tab:dynamic3d_domains}
\renewcommand{\arraystretch}{1.2}

\begin{adjustbox}{width=\textwidth}
\begin{tabular}{
  >{\centering\arraybackslash}m{0.16\textwidth}
  m{0.36\textwidth}
  m{0.24\textwidth}
  m{0.12\textwidth}
  >{\centering\arraybackslash}m{0.12\textwidth}
}
\toprule[1.5pt]
\textbf{Domain} &
\textbf{Description} &
\textbf{Core Challenges} &
\textbf{Variants} &
\textbf{State / Action} \\
\midrule

\includegraphics[width=0.14\textwidth]{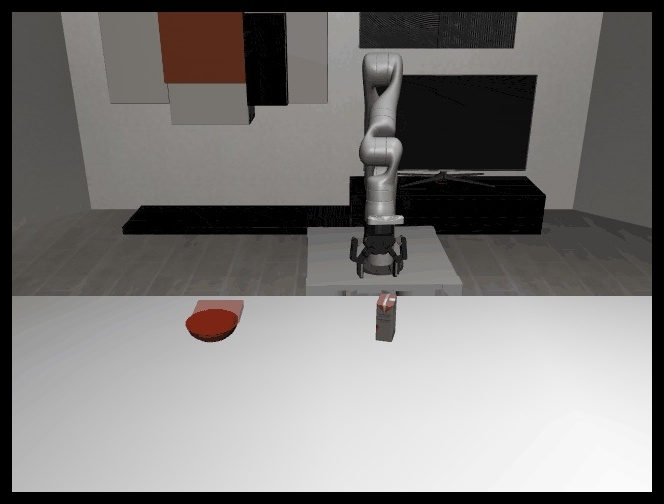}
&
\textbf{Rearrange3D.}
A 3D task where the robot must rearrange objects into different spatial arrangements with respect to other objects.

&
\makecell[l]{
\textbf{Dynamic3D} \\
Basic Spatial Relations
}
&
Each variant requires the robot to put one or more objects on the left, right, front, behind, or next to another object.
&
\makecell[c]{
$\mathcal{S} \in \mathbb{R}^{N \times 16 + 57}$ \\
$\mathcal{A} \in \mathbb{R}^{11}$
} \\

\midrule

\includegraphics[width=0.14\textwidth]{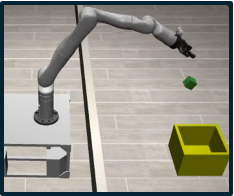}
&
\textbf{Tossing3D.}
A 3D task where an object that is initially on the floor must be transferred to the bin. The robot must toss the object into a bin, since it cannot reach the goal position due to an immovable obstacle.
&
\makecell[l]{
\textbf{Dynamic3D} \\
Dynamic Constraints
}
&
The variants require tossing different numbers of objects into the bin.
&
\makecell[c]{
$\mathcal{S} \in \mathbb{R}^{N \times 16 + 54}$ \\
$\mathcal{A} \in \mathbb{R}^{11}$
} \\

\midrule

\includegraphics[width=0.14\textwidth]{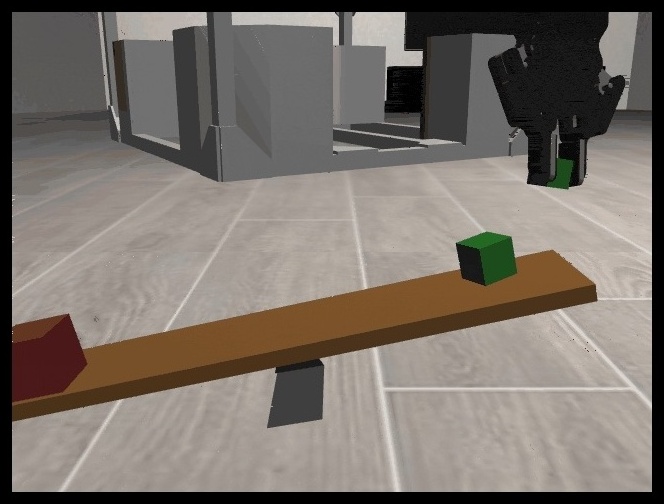}
&
\textbf{BalanceBeam3D.}
A 3D task where the robot must balance multiple objects onto a balance beam, which requires understanding of spatial relationships between objects of different sizes, and the rotational forces they induce.
&
\makecell[l]{
\textbf{Dynamic3D} \\
Dynamic Constraints \\
Tool Use
}
&
This task has one variant with three objects to balance.
&
\makecell[c]{
$\mathcal{S} \in \mathbb{R}^{N \times 16 + 38}$ \\
$\mathcal{A} \in \mathbb{R}^{11}$
} \\

\midrule

\includegraphics[width=0.14\textwidth]{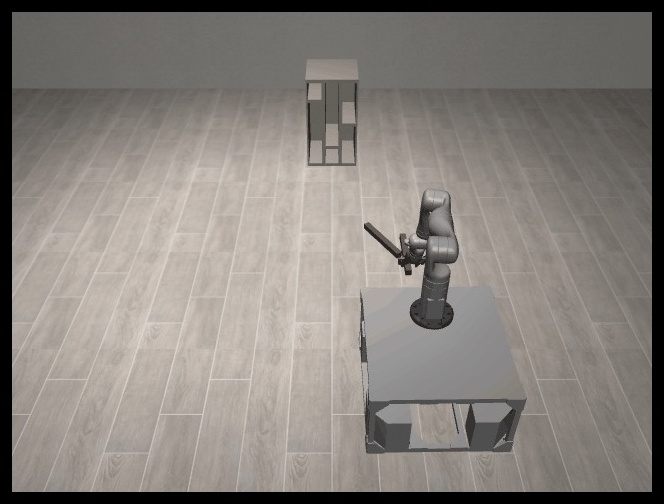}
&
\textbf{ConstrainedCupboard3D.}
A 3D task where the robot is supposed to fit multiple long rods into constrained spaces in a cupboard. The cupboard has varying numbers and sizes of rows and columns.
&
\makecell[l]{
\textbf{Dynamic3D} \\
Combinatorial Geometric Constraints
}
&
The variants require fitting a different number of objects into cupboards of different sizes with varying arrangements of feasible regions at each reset.

&
\makecell[c]{
$\mathcal{S} \in \mathbb{R}^{N \times 16 + M \times 6 + 22}$ \\
$\mathcal{A} \in \mathbb{R}^{11}$
} \\

\midrule
\includegraphics[width=0.14\textwidth]{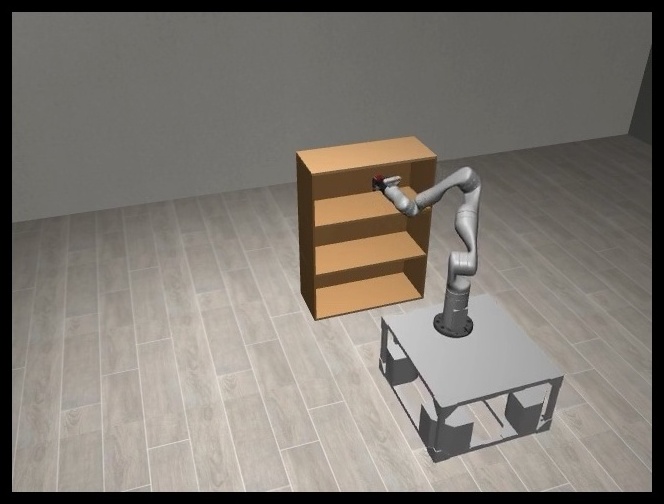}
&
\textbf{Shelf3D.}
A 3D task where the robot must pick up objects from the ground and place them onto a space-constrained shelf in a cupboard with three layers.
&
\makecell[l]{
\textbf{Dynamic3D} \\
Basic Spatial Relations \\
Combinatorial Geometric Constraints
}
&
The variants require picking and placing different numbers of objects.
&
\makecell[c]{
$\mathcal{S} \in \mathbb{R}^{N \times 12 + 26}$ \\
$\mathcal{A} \in \mathbb{R}^{11}$
} \\

\midrule

\includegraphics[width=0.14\textwidth]{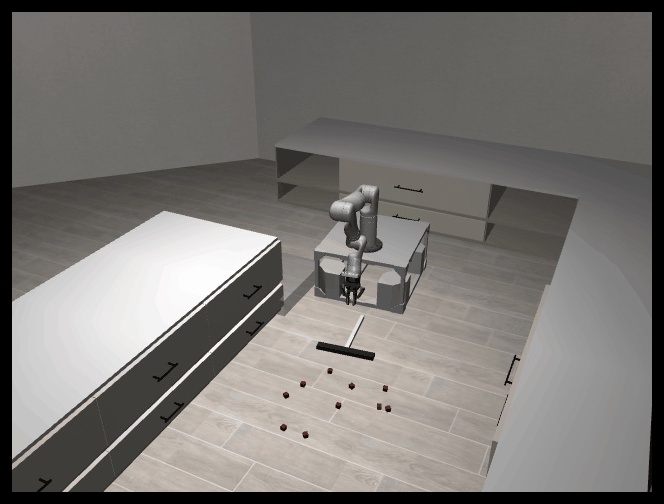}
&
\textbf{SweepSimple3D.}
A 3D task where the robot must sweep objects that are spread out on the floor into different regions around fixtures in the room. A broom tool is available that may be used for sweeping.
&
\makecell[l]{
\textbf{Dynamic3D} \\
Tool Use \\
Nonprehensile Multi-Object Manip.
}
&
These variants have different numbers of objects to sweep and multiple goal locations.
&
\makecell[c]{
$\mathcal{S} \in \mathbb{R}^{N \times 16 + 73}$ \\
$\mathcal{A} \in \mathbb{R}^{11}$
} \\

\midrule

\includegraphics[width=0.14\textwidth]{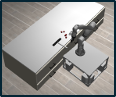}
&
\textbf{SweepIntoDrawer3D.}
A 3D task where the robot must open a drawer and sweep a pile of objects into the drawer. A brush tool is available that may be used for sweeping.
&
\makecell[l]{
\textbf{Dynamic3D} \\
Tool Use \\
Nonprehensile Multi-Object Manip. \\
Dynamic Constraints.
}
&
This task has one variant requiring a fixed number of objects to be swept into the drawer.
&
\makecell[c]{
$\mathcal{S} \in \mathbb{R}^{N \times 16 + 73}$ \\
$\mathcal{A} \in \mathbb{R}^{11}$
} \\

\midrule

\includegraphics[width=0.14\textwidth]{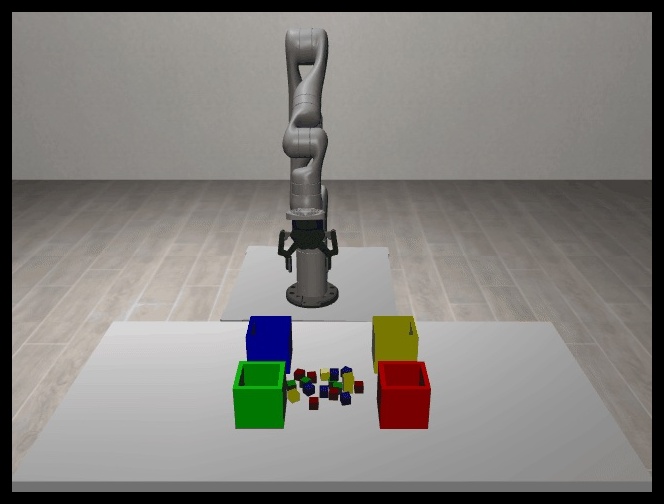}
&
\textbf{SortClutteredBlocks3D.}
A 3D task where the robot must sort a pile of objects into different receptacles based on their color. The objects may be initially in contact with each other, requiring singulation before grasping.
&
\makecell[l]{
\textbf{Dynamic3D} \\
Basic Spatial Relations \\
Nonprehensile Multi-Object Manip. \\
}
&
The variants differ in the number of objects to sort and in the types of receptacles (a cupboard or bins).
&
\makecell[c]{
$\mathcal{S} \in \mathbb{R}^{N \times 16 + M \times 16 + 29}$ \\
$\mathcal{A} \in \mathbb{R}^{11}$
} \\

\midrule

\includegraphics[width=0.14\textwidth]{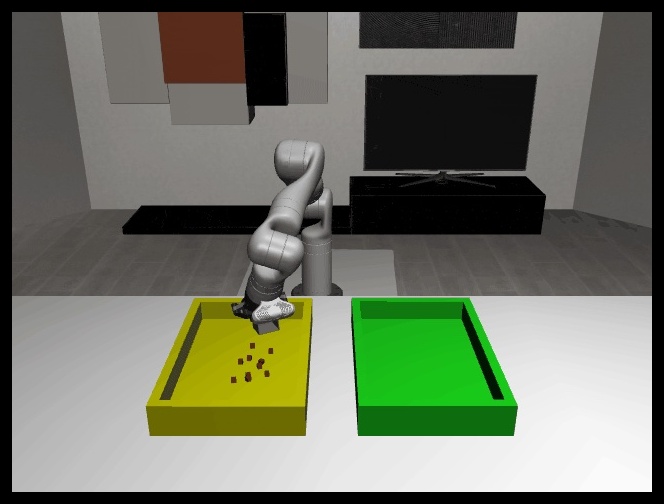}
&
\textbf{ScoopPour3D.}
A 3D task where the robot must transfer a pile of objects from one bin to another. There is a tool available that may be used for scooping and pouring.
&
\makecell[l]{
\textbf{Dynamic3D} \\
Tool Use \\
Nonprehensile Multi-Object Manip. \\
Dynamic Constraints.
}
&
The variants require scooping and pouring different numbers of objects.
&
\makecell[c]{
$\mathcal{S} \in \mathbb{R}^{N \times 16 + 105}$ \\
$\mathcal{A} \in \mathbb{R}^{11}$
} \\

\bottomrule[1.5pt]
\end{tabular}
\end{adjustbox}
\end{table*}

We present detailed descriptions of the environments developed in \gardenname{} in Tables~\ref{tab:kinematic2d_domains}, \ref{tab:dynamic2d_domains}, \ref{tab:kinematic3d_domains}, and \ref{tab:dynamic3d_domains}.
Different variants within the same environment may have different state spaces, depending on the number of objects ($N$); we report the state and action spaces in the corresponding tables.

\subsection{\gymname{} Additional Details}
\label{app:kindergym-additional-details}
\myparagraph{Teleoperation:}
We used three devices to collect demonstrations. For the 2D environments, we use either a PS5 controller or a keyboard to directly control the action space. 
For the 3D environments, users operate in three distinct modes that separately control the base, arm, and gripper. 
For the base and gripper, we directly command joint values, whereas for the arm we specify the end-effector pose and compute the corresponding joint values using an inverse kinematics  solver.

\subsection{\benchmarkname{} Additional Details}
\label{app:bench-additional-details}
\myparagraph{Implementation Details about Baselines:}
\begin{enumerate}
    \item \textbf{Bilevel Planning~(BP)}~\cite{srivastava2014combined,silver2023predicateinvent,li2025IVNTR}: We use search-then-sample bilevel planning. Abstract plans are generated with greedy best-first search using the hFF heuristic. We consider a maximum of 10 abstract plans or a wall-clock timeout of 60 seconds, whichever comes first. Each abstract plan is refined with a backtracking search over samplers, with a maximum of 3 samples per step.
    \item \textbf{LLM Planning (LLMPlan)}~\cite{silver2022pddl,song2023llm}: A zero-shot planner that prompts an LLM (GPT-5.2~\cite{singh2025openai}) with available skills, object-centric initial state, and goal description. LLM output is parsed to obtain an ordered list of skills and their parameters, which are rolled-out open-loop. We provide the prompt template in Figure~\ref{fig:llmplan-prompt}.
    \item \textbf{VLM Planning (VLMPlan)}~\cite{hu2023look}: A zero-shot VLM (GPT-5.2~\cite{singh2025openai}) planner that uses \emph{RGB images} of the initial environment state, along with states and skills. We use the same prompt template as in LLMPlan.
    \item \textbf{LLM In-context (LLMCon)}~\cite{silver2022pddl,song2023llm}: Similar to LLMPlan, this is also an LLM (GPT-5.2~\cite{singh2025openai}) planner that is prompted with skill options, object-centric initial state, and goal description, but with additional in-context examples (up to 4 in count, depending on task). We provide the prompt template in Figure~\ref{fig:llmpcon-prompt}.
    \item \textbf{VLM In-context (VLMCon)}~\cite{hu2023look}: A VLM (GPT-5.2~\cite{singh2025openai}) planner that is prompted with states and skills, in-context examples, and additionally \emph{RGB images} of the initial state of the environment. We use the same prompt as in LLMCon.
    \item \textbf{Model Predictive Control~(MPC)}~\cite{howell2022predictive}: We perform predictive sampling trajectory optimization.
    For each iteration, we sample 10 candidate action sequences, each of horizon 100, by adding Gaussian noise to the current best trajectory (warm-started from the previous iteration). 
    Then we rollout each candidate trajectory using the ground-truth simulation transition functions and score each trajectory using the sparse reward function. 
    The first action of the lowest-cost candidate will be executed and start the next iteration. 
    We use 10 control points to ensure smooth trajectories and reduce the effective search dimensionality. 
    \item \textbf{Model-based Reinforcement Learning~(MBRL)}: We use the demonstrations collected for imitation learning baselines to train a neural (state-based) transition model~\cite{chitnis2022learning}. 
    The neural transition model takes the input as the current state and action and predicts the delta change of the robot state and environment states. 
    The neural transition functions contain two hidden layers, each layer with 256 neurons with SiLU activation function~\cite{elfwing2018sigmoid}. 
    The model contains two linear output heads for predicting the change of robot states and environment states. 
    Once the neural transition model is learned, we run MPC to select the best actions with the same sparse reward function. 
    \item \textbf{Generative Diffusion Planning}: Following the official implementation of Generative Skill Chaining (GSC)~\cite{mishra2023gsc}, we first convert our demonstration trajectories to the Ant-Maze data format. 
    We then train the diffusion-based framework that models each skill as a probabilistic distribution and composes them at inference time to generate entire plans while enabling scalable, constraint-aware planning for unseen tasks.
    \item \textbf{Proximal Policy Optimization (PPO)}~\cite{schulman2017proximal}:
    We use the standard PPO actor-critic architecture, where both the actor and the critic are parameterized by multilayer perceptrons (MLPs).
    Each MLP consists of four hidden layers.
    For 2D environments, the hidden layer dimensions are $[128, 128, 128, 128]$, while for 3D environments they are $[256, 256, 256, 256]$.
    All hidden layers use \texttt{tanh} activations.
    The PPO clip ratio is set to $0.2$ for all environments.
    Agents are trained for a total of $1$M environment steps.
    Both the actor and critic are optimized using Adam with a learning rate of $3 \times 10^{-4}$.
    
    \item \textbf{Soft Actor-Critic (SAC)}~\cite{haarnoja2018soft}:
    We follow a similar network architecture as PPO.
    Both the actor and the critic (Q-value estimator) are implemented as MLPs with four hidden layers.
    For 2D environments, the hidden layer dimensions are $[128, 128, 128, 128]$, and for 3D environments they are $[256, 256, 256, 256]$.
    All hidden layers use \texttt{tanh} activations.
    We use a replay buffer of size $500{,}000$ and delay learning until $5{,}000$ environment steps have been collected.
    The policy and Q-value networks are optimized using Adam, with learning rates of $3 \times 10^{-4}$ and $1 \times 10^{-3}$, respectively.
    Agents are trained for a total of $1$M environment steps.
    \item \textbf{Diffusion Policy (DP)}~\cite{chi2025diffusion}: We employ a conditional diffusion policy with a hybrid image–state UNet backbone. The policy predicts action trajectories over a horizon of 16 steps, conditioning on the most recent 2 observation steps and executing 8 future action steps. The denoising network is a temporal UNet with channel dimensions [256, 512, 1024], kernel size 5, and Group Normalization with 8 groups. 
    A 128-dimensional diffusion timestep embedding is injected into all UNet blocks. Observations are encoded and used as global conditioning: visual inputs are processed by an image encoder with Group Normalization, and low-dimensional robot states are embedded via MLPs. We adopt a DDIM scheduler with 100 training timesteps, a squared cosine noise schedule, and epsilon prediction, using 16 denoising steps during inference.
    Models are trained using AdamW with a learning rate as ($10^{-4}$).
    \item \textbf{DP + Environment States (DPES)}~\cite{chi2025diffusion}: This method extends Diffusion Policy by incorporating additional low-level environment states as input. These environmental states are encoded using MLPs. 
    \item \textbf{Finetuned VLA}~\cite{intelligence2025pi_}: we fine-tune $\pi_{0.5}$ using the same demonstrations as the diffusion policy using their open-sourced OpenPI codebase \cite{intelligence2025pi_}. We use the same state and action spaces as DP. Notably, the VLA model does not have access to environment states during either training or inference. We use action horizon of 10, and during inference we execute the first 8 actions. 
\end{enumerate}

\begin{figure*}[t]
\centering
\begin{tcolorbox}[
    colback=gray!10,     %
    colframe=black,      %
    title={LLMPlan prompt template},
    width=1.0\linewidth,
]
\begin{Verbatim}[breaklines=true,breaksymbolleft={},breaksymbolright={}]
You are highly skilled in robotic task planning, breaking down intricate and long-term tasks into distinct primitive actions.
Consider the following skills a robotic agent can perform. Note that each of these skills takes the form of a `ParameterizedController` and may have both discrete arguments (indicated by the `types` field, referring to objects of particular types),
as well as continuous arguments (indicated by `params_space` field, which is formatted as `Box([<param1_lower_bound>, <param2_lower_bound>, ...], [<param1_upper_bound>, <param2_upper_bound>, ...], (<number_of_params>,), <datatype_of_all_params>)`).

{controllers}

You are only allowed to use the provided skills. It's essential to stick to the format of these basic skills. When creating a plan, replace
the arguments of each skill with specific items or continuous parameters. You can first describe the provided scene and what it indicates about the provided
task objects to help you come up with a plan.

Here is a list of objects present in this scene for this task, along with their type (formatted as <object_name>: <type_name>):
{typed_objects}

And here are the available types (formatted in PDDL style as `<type_name1> <type_name2>... - <parent_type_name>`). You can infer a hierarchy of types via this:
{type_hierarchy}

Finally, here is an expression corresponding to the current task goal that must be achieved:
{goal_str}

Please return a plan that achieves the provided goal from an initial state described below.
{init_state_str}

Please provide your output in the following format (excluding the angle brackets and ellipsis, which are just for illustration purposes).
Be sure to include the parens '(' and ')', as well as square brackets '[' and ']' even if there are no objects/continuous parameters.
Do not bold or italicize or otherwise apply any extra formating to the plan text. Do not provide any numbers for steps in the plan, or
any reasoning for each step below the 'Plan:' heading:
<Explanation of scene + your reasoning>
Plan:
<skill 1 name>(<obj1_name>:<obj1_type_name>, <obj2_name>:<obj2_type_name>, ...)[<continuous_param1_value>, <continuous_param2_value>, ...]
<skill 2 name>(<obj1_name>:<obj1_type_name>, <obj2_name>:<obj2_type_name>, ...)[<continuous_param1_value>, <continuous_param2_value>, ...]
...
\end{Verbatim}
\end{tcolorbox}
\caption{Prompt template for the LLMPlan planner baseline. Strings in braces are replaced with task-specific content.}
\label{fig:llmplan-prompt}
\end{figure*}

\begin{figure*}[t]
\centering
\begin{tcolorbox}[
    colback=gray!10,     %
    colframe=black,      %
    title={LLMCon prompt template},
    width=1.0\linewidth,
]
\begin{Verbatim}[breaklines=true,breaksymbolleft={},breaksymbolright={}]
Create a high-level plan for completing a task using the allowed actions and visible objects.

Allowed actions: Consider the following skills a robotic agent can perform. Note that each of these skills takes the form of a `ParameterizedController` and may have both discrete arguments (indicated by the `types` field, referring to objects of particular types),
as well as continuous arguments (indicated by `params_space` field, which is formatted as `Box([<param1_lower_bound>, <param2_lower_bound>, ...], [<param1_upper_bound>, <param2_upper_bound>, ...], (<number_of_params>,), <datatype_of_all_params>)`).
{controllers}

Visible objects: Here is a list of objects present in this scene for this task, along with their type (formatted as <object_name>: <type_name>):
{typed_objects}

Here are the available types (formatted in PDDL style as `<type_name1> <type_name2>... - <parent_type_name>`). You can infer a hierarchy of types via this:
{type_hierarchy}

Task description:
{goal_str}

In-context examples:
{in_context_examples}

Please return a plan that achieves the provided goal from an initial state described below.
{init_state_str}

Completed plans: No plan has yet been executed.

Next plans: Please provide your output in the following format (excluding the angle brackets and ellipsis, which are just for illustration purposes).
Be sure to include the parens '(' and ')', as well as square brackets '[' and ']' even if there are no objects/continuous parameters.
Do not bold or italicize or otherwise apply any extra formating to the plan text. Do not provide any numbers for steps in the plan, or
any reasoning for each step below the 'Plan:' heading:
<Explanation of scene + your reasoning>
Plan:
<skill 1 name>(<obj1_name>:<obj1_type_name>, <obj2_name>:<obj2_type_name>, ...)[<continuous_param1_value>, <continuous_param2_value>, ...]
<skill 2 name>(<obj1_name>:<obj1_type_name>, <obj2_name>:<obj2_type_name>, ...)[<continuous_param1_value>, <continuous_param2_value>, ...]
...
\end{Verbatim}
\end{tcolorbox}
\caption{Prompt template for the LLMCon planner baseline. Strings in braces are replaced with task-specific content.}
\label{fig:llmpcon-prompt}
\end{figure*}

\myparagraph{Standard Deviations:} We present the standard deviations of the empirical results in Table~\ref{tab:emp_std}.

\myparagraph{Dense Reward for RL:} In addition to the default sparse reward setting, we also experiment with a dense reward formulation for RL baselines in the BaseMotion3D environment.
Specifically, we engineer a distance-based shaping reward that measures the change in Euclidean distance to the goal projected onto the XOY plane, and include this signal as a step-wise reward for SAC and PPO.
We report the performance comparison in Table~\ref{tab:rl_rwd} over 5 random seeds.
Without this carefully engineered dense reward, the performance of PPO baselines degrades substantially, highlighting the sensitivity of standard PPO to reward design in long-horizon, sparse-reward settings.
As an off-policy algorithm, SAC relies less on carefully engineered dense rewards and the same reward decreased performance.

\subsection{Additional Real-to-Sim-to-Real Details}
\label{app:real-to-sim-to-real}
We first detect object-centric states, including robot states and environment states. 
For robot states, we estimate the robot base pose in world coordinates using calibrated top-down cameras. 
We directly obtain proprioceptive states, such as arm joint angles and gripper state, from the robot. 
For environment object-centric states, we detect 2D bounding boxes for each object using~\cite{gu2021open}. 
We use the center of each bounding box to estimate object position and the bounding box size to represent object scale, while keeping the z-axis dimension fixed in simulation. 
Once the simulation is constructed, we generate task and motion plans using our bilevel planning baseline. 
We then execute the generated plans on the real-world robot. 
During execution, we ensure that the real robot sequentially achieves each waypoint in the plan.
We expect these initial experiments to open future directions for building real-to-sim-to-real systems for more complex tasks.

\subsection{Noisy Observations and Actions}
To capture the challenge of perception noise in real-world scenarios, we implemented observation and action wrappers that add noise to KinDER observations and actions to simulate simple stochasticity in perception and dynamics.
In Table~\ref{tab:noise}, we show bilevel planning with varying noise levels.

\end{document}